\documentclass{article}

\usepackage{graphicx}
\usepackage{soul}
\usepackage[preprint]{neurips_2026}

\usepackage[utf8]{inputenc} 
\usepackage[T1]{fontenc}    
\usepackage{hyperref}       
\usepackage{url}            
\usepackage{booktabs}       
\usepackage{amsfonts}       
\usepackage{nicefrac}       
\usepackage{microtype}      
\usepackage[dvipsnames]{xcolor}         
\usepackage{amsmath}
\usepackage{booktabs}
\usepackage{multirow}
\usepackage{titlesec}
\usepackage{titletoc}
\usepackage{wrapfig}

\definecolor{copper}{rgb}{0.82, 0.38, 0.10}
\definecolor{niceblue4}{rgb}{0.08, 0.60, 0.75}
\definecolor{softrose}{rgb}{0.90, 0.40, 0.55}
\hypersetup{
    colorlinks,
    linkcolor=copper,
    citecolor=niceblue4,
    urlcolor={blue}
}

\definecolor{pastelBlue}{RGB}{198, 235, 245}   
\definecolor{pastelGreen}{RGB}{248, 214, 210}  

\newcommand{\hlSmall}[1]{\sethlcolor{pastelBlue}\hl{#1}}
\newcommand{\hlMedium}[1]{\sethlcolor{pastelGreen}\hl{#1}}

\title{Representation Learning Enables Scalable \\ Multitask Deep Reinforcement Learning}

\author{%
Johan Obando-Ceron\textsuperscript{1,2}\quad
Lu Li\textsuperscript{1,2}\quad
Scott Fujimoto\textsuperscript{3}\quad
Pierre-Luc Bacon\textsuperscript{1,2}\\[0.1em]
\textbf{Aaron Courville\textsuperscript{1,2,4}}\quad
\textbf{Pablo Samuel Castro\textsuperscript{1,2,5}}\\[1em]
\textsuperscript{1}Mila -- Qu\'ebec AI Institute\quad
\textsuperscript{2}Universit\'e de Montr\'eal\quad
\textsuperscript{3}McGill University\quad\\
\textsuperscript{4}CIFAR AI Chair\quad
\textsuperscript{5}Google DeepMind\\[1em]
\texttt{{jobando0730}@gmail.com}, \texttt{scott.fujimoto@mail.mcgill.ca}\\[4pt]
\texttt{\{lu.li, pierre-luc.bacon, courvila, pablo-samuel.castro\}@mila.quebec}
}

\begin{document}

\maketitle

\begin{abstract}
Scaling reinforcement learning (RL) to diverse multitask settings remains a central challenge. While recent advances in model-based RL achieve strong performance, they rely on planning and complex training pipelines, making it unclear which components are essential for scalability. We revisit this question and argue that the primary driver of scalable multitask RL is not model-based control, but \emph{representation learning}. In particular, we show that combining predictive, model-based representations with high-capacity value function approximation is sufficient to achieve strong performance, even without planning. We evaluate a simple model-free algorithm, MR.Q, coupled with auxiliary predictive objectives into a scalable actor-critic architecture. This approach outperforms a recent world-model-based method and a range of deep RL baselines across a diverse suite of multitask continuous control tasks, while significantly reducing computational overhead and improving wall-clock efficiency. We observe consistent improvements with increased model capacity and show through ablations that predictive representation learning is critical for performance.
\textbf{Our code is available at \href{https://github.com/johanobandoc/ScaleMRL.git}{ScaleMRL}.}

\end{abstract}

\begin{center}
\begin{minipage}{0.84\linewidth}
\itshape
``What we observe isn't nature itself, but nature exposed to our method of questioning\footnotemark.''

\hfill --- Werner Heisenberg
\end{minipage}
\end{center}

\footnotetext{In RL, what an agent “sees” depends on its representation. Our results suggest that improving representations can be more important, and significantly more efficient, than modeling environment dynamics and planning.}

\section{Introduction}

Deep reinforcement learning (RL) has achieved remarkable success across a wide range of domains, including games, robotics, and control~\citep{akkaya2019solving,mnih2013playing,schwarzer2023bigger}. However, much of this progress remains confined to single-task settings, where agents are trained and evaluated on narrowly defined environments, often requiring hundreds of millions of environment interactions to converge. In contrast, recent advances in machine learning, particularly in language and vision, demonstrate that scaling models across diverse data distributions enables generalization, transfer, and robustness through shared representations~\citep{wang22u,flamingo_2022,kojima2022large,subramanian2023towards,zhou2025weak,reed2022a,wiedemer2026video}. Extending these principles to online deep RL remains an open challenge. Unlike supervised settings, RL involves non-stationary data, bootstrapped targets, and long-horizon credit assignment, which introduce optimization instabilities that manifest as representation collapse, loss of plasticity, and unstable value estimation. These instabilities compound the sample costs of learning and ultimately hinder progress in multitask settings~\citep{kumar2021implicit,nikishin2022primacy,sokar2023dormant,nauman2024overestimation,tang2024improving,castanyer2025stable}.

Multitask RL (MTRL) seeks to train a single agent over a distribution of tasks, but doing so across increasingly diverse task distributions introduces instability, task interference, and underutilization of model capacity~\citep{teh2017distral,yu2020gradient,Eramo2020Sharing,kong2025mastering}. Recent work by \citet{nauman2025bigger} demonstrates that substantially increasing value function capacity, paired with categorical value parameterization and explicit regularization, leads to significant multitask gains. Yet scaling model size alone does not solve the problem: without the right training objectives and representation learning mechanisms, larger models simply require more data to stabilize~\citep{taiga2023investigating,farebrother24a}. This points to representation quality as a central axis of progress, since better representations have been shown to reduce TD variance, accelerate learning, and stabilize training across tasks~\citep{castro2021mico,schwarzerdata,fujimoto2023sale,cetin2023hyperbolic,echchahed2025a,ceron2026simplicial}.

Model-based RL methods pursue this goal by leveraging predictive objectives — specifically by learning latent dynamics models — to provide dense supervision that shapes representations beyond what TD learning alone can achieve. This richer learning signal is a key driver behind recent model-based advances~\citep{hafner2020mastering, hafner2023mastering, hansen2023td, Hansen2025Newt, fujimototowards}. Recent large-scale systems further combine predictive representation learning, large shared architectures, and planning to achieve strong multitask performance~\citep{xu2023on,georgiev2024pwm,hafner2023mastering,Hansen2025Newt}. Yet because these approaches bundle multiple components together, isolating the source of their gains remains difficult. Moreover, planning itself introduces computational overhead, hyperparameter sensitivity, and compounding model errors, ultimately eroding the very efficiency gains these methods aim to provide~\citep{zhang21n,erikmodelreg,rajeswaran2017epopt,clavera18a,chua2018deep,voelcker2022value}.

We hypothesize that much of the benefit attributed to model-based control in fact arises from the representations these methods learn, and that predictive objectives alone are sufficient to achieve competitive sample efficiency at scale~\citep{jaderberg2017reinforcement,gelada2019deepmdp,lee2020predictive,anand2022procedural}. To test this hypothesis, we study MR.Q~\citep{fujimototowards}, a purely model-free agent that integrates predictive objectives into TD learning. MR.Q is a natural probe for this question as it isolates the representational benefits of predictive learning from the confounds of planning, allowing us to test whether richer supervision alone drives sample efficiency gains.

While originally proposed for single-task settings, we extend MR.Q's evaluation to the multitask regime. However, previous MTRL benchmarks evaluate at 100M or more environment steps~\citep{Hansen2025Newt}, obscuring whether methods are genuinely sample-efficient or simply benefit from prolonged training. To address this, we consider a version of the benchmark that evaluates agents at 10M environment steps, where sample efficiency gains are most visible.

Across a suite of continuous control benchmarks, MR.Q outperforms a recent world-model-based method (\textit{Newt}~\citep{Hansen2025Newt}) while achieving substantially improved wall-clock, sample efficiency, and demonstrates performance benefits from scaling in both model size and data availability. In addition, MR.Q exhibits stronger transfer to unseen tasks than \textit{Newt}, suggesting that representations learned through multitask training yield substantially better zero-shot initialization and faster adaptation during few-shot finetuning. Ablations further confirm that predictive objectives are critical, with performance degrading significantly when removed even at large model scales. Overall, these results support a representation-centric view of deep RL scaling, where the quality of learned representations plays a central role in enabling effective scalable multitask learning.

\section{Preliminaries}

\paragraph{Problem setting.}
We consider a multitask RL (MTRL) setting in which tasks $\tau \sim p(\tau)$ are sampled from a task distribution. Each task induces a Markov decision process (MDP) $\mathcal{M}_\tau = (\mathcal{S}, \mathcal{A}, \mathcal{T}_\tau, \mathcal{R}_\tau, \gamma)$, where we assume a shared action space $\mathcal{A}$ and (typically) a shared state space $\mathcal{S}$ across tasks, while transition dynamics and rewards may vary with $\tau$. At each time step $t$, the agent observes $s_t \in \mathcal{S}$, takes action $a_t \in \mathcal{A}$, and receives reward $r_t \sim \mathcal{R}_\tau(s_t, a_t)$, transitioning to $s_{t+1} \sim \mathcal{T}_\tau(\cdot \mid s_t, a_t)$. The objective is to learn a single policy $\pi(a \mid s, \tau)$ that maximizes the expected discounted return across tasks, formulated as $
\mathbb{E}_{\tau \sim p(\tau), \pi} \left[ \sum_{t=0}^{\infty} \gamma^t r_t \right].$  Similar to \citet{Hansen2025Newt}, when task information is available (e.g., task identifiers or language instructions), we condition the policy and value functions on a learned embedding $e(\tau)$. Otherwise, the problem reduces to a partially observable MDP, where task identity must be inferred from interaction. We assume an off-policy setting, where experience is stored in a replay buffer $\mathcal{D}$ containing tuples $(s_t, a_t, r_t, d_t, s_{t+1}, \tau)$, with $d_t \in \{0,1\}$ indicating episode termination. We adopt an off-policy actor--critic architecture~\citep{konda1999actor,fujimoto2018addressing}, where a parametric policy (actor) $\pi_\psi(a \mid s, \tau)$ is trained to maximize expected return, while a value function (critic) $Q_\theta(s,a,\tau)$ estimates the expected return of state--action pairs. The critic is optimized via temporal-difference (TD) learning using targets constructed from a slowly updated target network, while the actor is trained to maximize the critic’s value estimates. In practice, we employ twin critics $Q_{\theta_1}, Q_{\theta_2}$ to mitigate overestimation bias, as in prior work on off-policy RL~\citep{fujimoto2018addressing, haarnoja2018soft}.

\paragraph{Predictive Information Representations.}

Representation learning is central to deep RL, particularly in high-dimensional and multitask settings where stability and generalization depend on the structure of learned features~\citep{agarwal2021learning,echchahed2025a}. 
Because supervision from temporal-difference learning is often weak and non-stationary, predictive auxiliary objectives are commonly used to stabilize optimization and encourage latent representations to capture environment dynamics and temporal structure beyond reward signals~\citep{nikishin2022primacy,hafner2020mastering,hansen2023td}. We consider an off-policy actor--critic operating on learned latent representations. Observations (and optionally task information) are encoded as $z_t = \phi_\xi(s_t, \tau)$, and both the policy $\pi_\psi(a \mid z)$ and twin critics $Q_{\theta_1}, Q_{\theta_2}$ operate in latent space. Critics are trained via temporal-difference learning with target networks, while the policy maximizes value estimates. To improve representation quality, we augment training with predictive modeling in latent space: models of dynamics, reward, and termination predict $(z_{t+1}, r_t, d_t)$ from $(z_t, a_t)$ \citep{fujimototowards}, and their gradients are backpropagated through the encoder $\phi_\xi$. This encourages representations that are predictive of environment dynamics and task-relevant signals. Crucially, no planning is performed, the learned models are used solely to shape the representation, isolating the benefits of predictive learning without the computational overhead and instability of explicit model-based control.

\section{Scaling deep RL through Representation Learning}

\begin{wrapfigure}{r}{0.6\textwidth}
    \centering
    \includegraphics[width=0.6\textwidth]{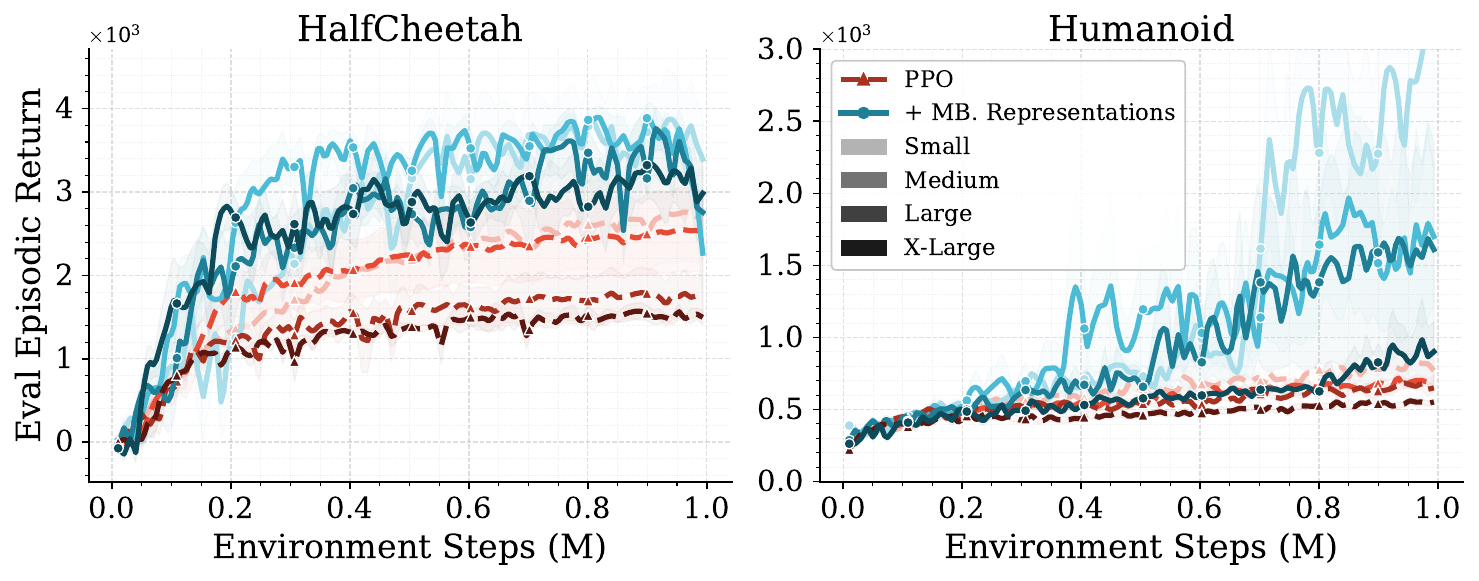}%
    \vspace{-0.4cm}
     \caption{\textbf{Representation quality drives scaling performance in model-free RL.} We compare standard PPO with a variant augmented with model-based representations (+\,MB.\ Representations) across four network sizes (Small, Medium, Large, X-Large) on HalfCheetah and Humanoid.}
    \vspace{-0.1cm}
    \label{fig:representation}
\end{wrapfigure}

A central challenge in deep RL is how to scale agents across tasks, model capacity, and data. Recent progress has been largely driven by model-based approaches, where agents learn predictive world models and leverage planning to improve decision-making \citep{hansen2023td,hafner2023mastering}. Methods such as Dreamer and TD-MPC2 demonstrate that combining predictive modeling with large-capacity function approximators can substantially improve performance in both single-task and multitask settings. At larger scales, systems such as \textit{Newt}~\citep{Hansen2025Newt} extend this paradigm to hundreds of tasks by training shared world models across diverse continuous-control domains, demonstrating strong multitask performance and transfer. However, these gains come with significant computational and algorithmic overhead. Model-based agents must jointly learn dynamics, reward, and value functions while additionally performing latent rollouts or planning during training or inference. This increases wall-clock cost, memory usage, and implementation complexity, while also introducing additional sources of instability as model errors compound over imagined trajectories \citep{erikmodelreg,jannerneurips}. These challenges become particularly pronounced in multitask settings, where a single world model must capture diverse and potentially conflicting dynamics across environments. Additional discussion of related work is provided in \autoref{appndx:related_work}.

At the same time, recent work suggests that some benefits commonly attributed to model-based RL may instead arise from the representations induced by predictive learning \citep{schwarzerdata,ghugare2023simplifying,zhao23k}. In particular, methods such as MR.Q show that model-free agents augmented with auxiliary predictive objectives can achieve strong performance across diverse tasks without explicit planning \citep{fujimototowards}. 

To isolate this effect, we study a controlled single-task setting where planning and multitask interference are absent. \autoref{fig:representation} provides evidence in a controlled setting where planning and multitask interference are absent. We compare standard PPO \citep{schulman2017proximal} with a variant augmented with predictive model-based representations (+\,MB.\ Representations) across four network sizes on HalfCheetah and Humanoid, two environments of increasing complexity and dimensionality. Without predictive representations, scaling model capacity yields little to no benefit: on HalfCheetah, larger PPO models can even underperform smaller ones, while on Humanoid, performance remains nearly flat across all model sizes. With predictive representations, however, PPO consistently outperforms standard PPO at every network size and offers increased robustness to varied capacity. These results hint that representation quality may be an important bottleneck when scaling deep RL systems. Predictive objectives provide an additional supervisory signal that appears to help larger models make more effective use of increased capacity, whereas reward-only supervision often struggles to do so.

This finding has direct implications for multitask RL, where scaling shared architectures is critical for learning transferable representations across diverse tasks~\citep{nauman2025bigger,Hansen2025Newt}, and where additional challenges — task interference, non-stationarity, and distributional shift — may make representation quality an even more severe bottleneck. This motivates the central question of this work: \emph{Can model-free RL match the scalability and generalization of world-model approaches in multitask settings by focusing on representation learning alone?}

\section{Multitask Model-Free RL with Structured Representations}
\label{sec:multitask_rl}

In this section, we evaluate whether model-free RL augmented with predictive representation learning can match recent world-model approaches in multitask settings. We show that MR.Q consistently matches or surpasses the large-scale world-model baseline \textit{Newt} across diverse multitask domains without relying on planning or latent rollouts. We further analyze how predictive representation learning impacts representation geometry and optimization stability.

World models provide useful inductive biases through predictive supervision and structured latent representations \citep{ha2018world,Hafner2020Dream,gelada2019deepmdp,schwarzerdata}. However, many of their benefits may arise from the learned representations rather than planning itself. This motivates model-free approaches that incorporate predictive representation learning while preserving the simplicity, efficiency and scalability of model-free RL.

\paragraph{Baselines and Evaluation Protocol.}
We compare against a strong model-based baseline, \textit{Newt} \citep{Hansen2025Newt}, in a multitask setting under fixed interaction budgets. Our primary evaluation is conducted in a low-data regime of 10M environment steps, where sample efficiency is critical, in contrast to prior work that typically evaluates at 100M environment steps~\citep{Hansen2025Newt}. To assess scalability, we additionally include selected longer runs. We report aggregate learning curves to evaluate sample efficiency, as well as final performance at the end of training, averaging results over five seeds and reporting 95\% confidence intervals (CIs) across tasks and runs. Our results show that equipping TD3~\citep{fujimoto2018addressing} with predictive representation learning objectives (MR.Q~\citep{fujimototowards}) enables model-free methods to match or surpass model-based approaches. All experiments follow the multitask language-conditioned training protocol introduced in \textit{Newt}~\citep{Hansen2025Newt}; see \autoref{appndx:mrq} and \autoref{appndx:protocol} for MR.Q and training details.

\begin{figure*}[!h]
    \centering
    \includegraphics[width=0.2\textwidth]{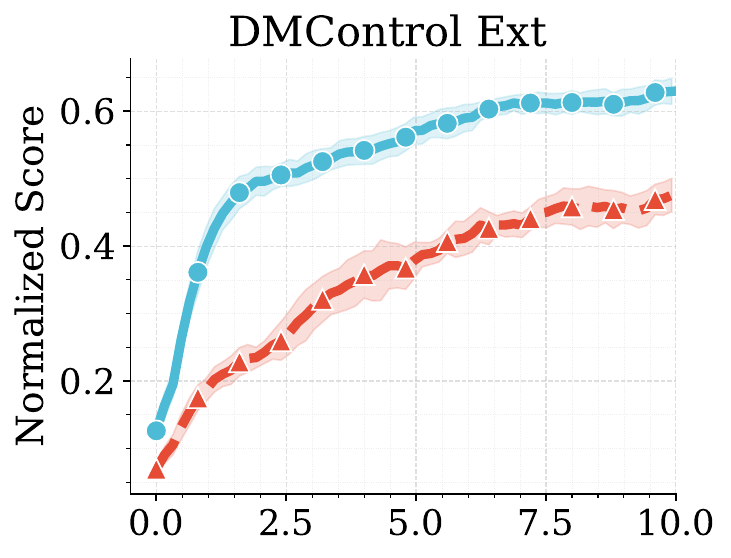}%
    \includegraphics[width=0.2\textwidth]{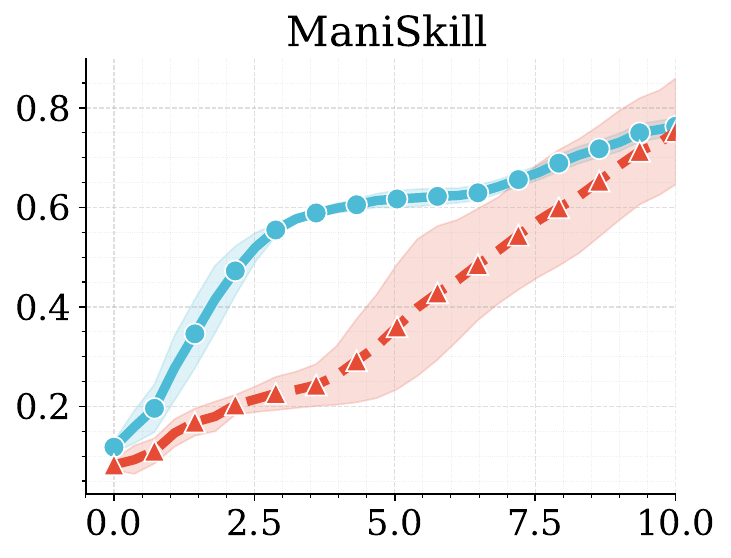}%
    \includegraphics[width=0.2\textwidth]{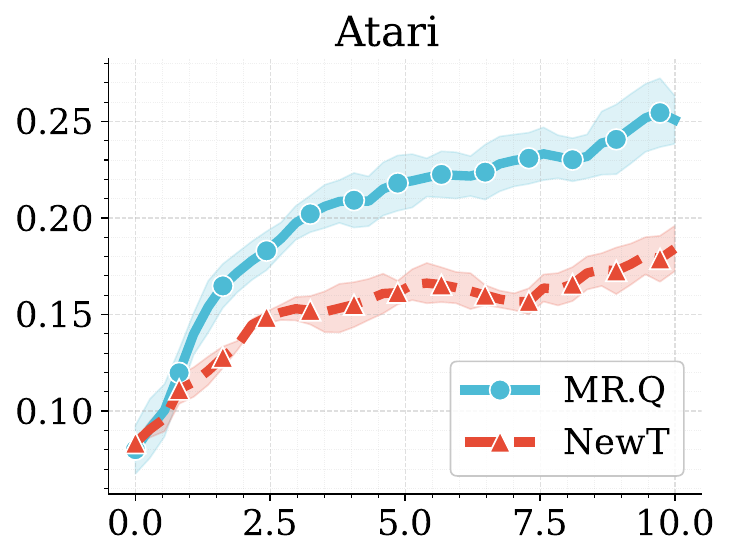}%
     \includegraphics[width=0.2\textwidth]{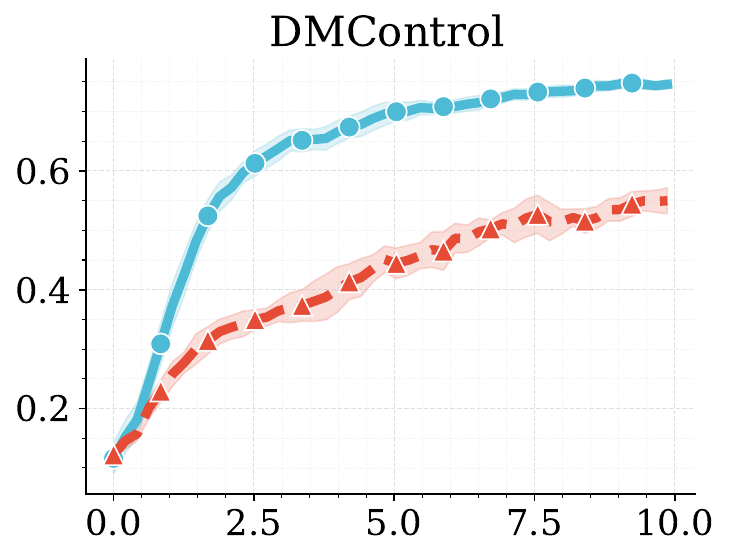}%
    \includegraphics[width=0.2\textwidth]
    {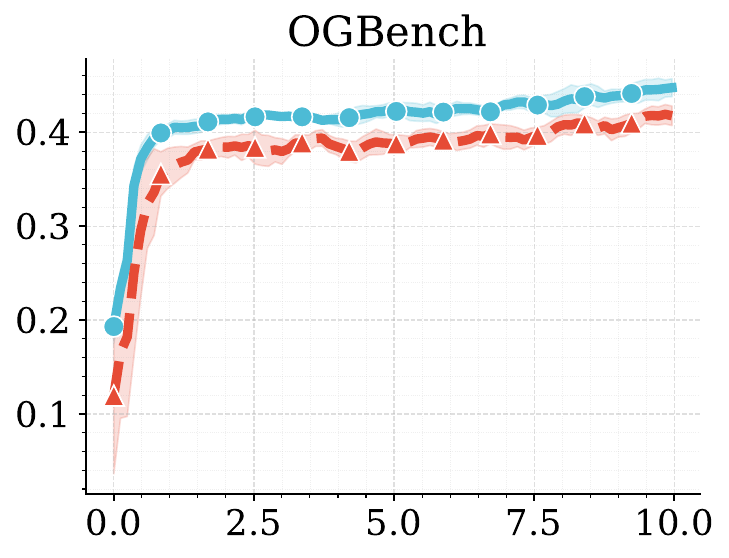}
    \includegraphics[width=0.2\textwidth]{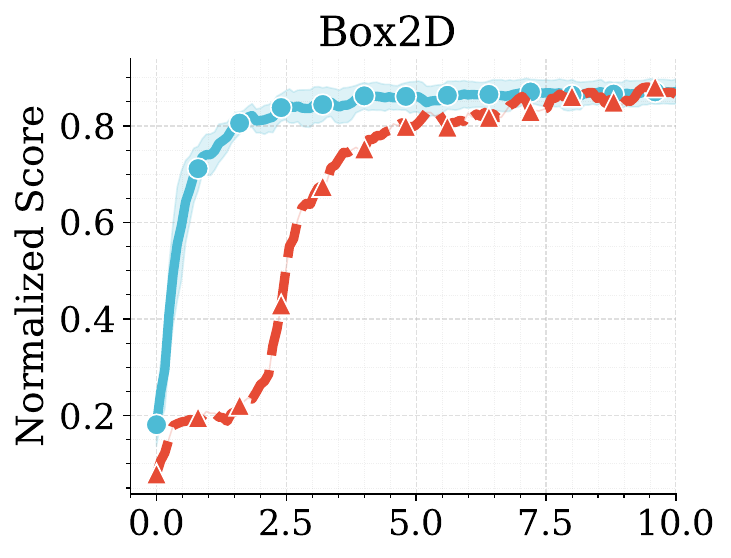}%
    \includegraphics[width=0.2\textwidth]{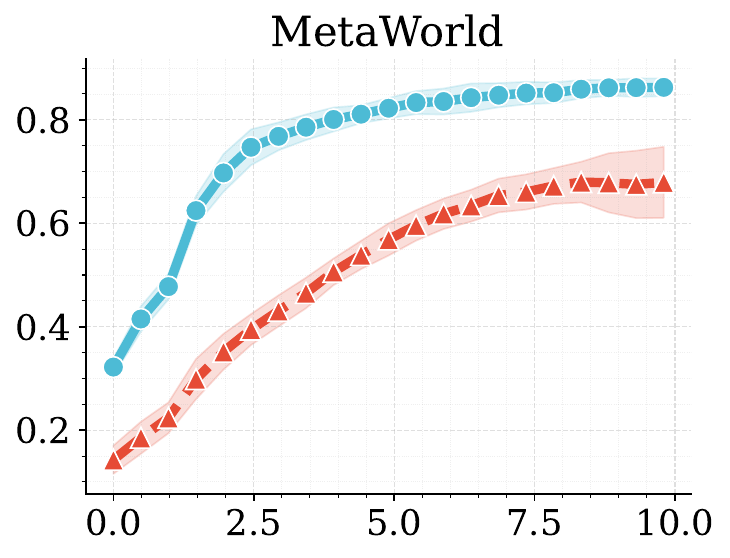}%
    \includegraphics[width=0.2\textwidth]{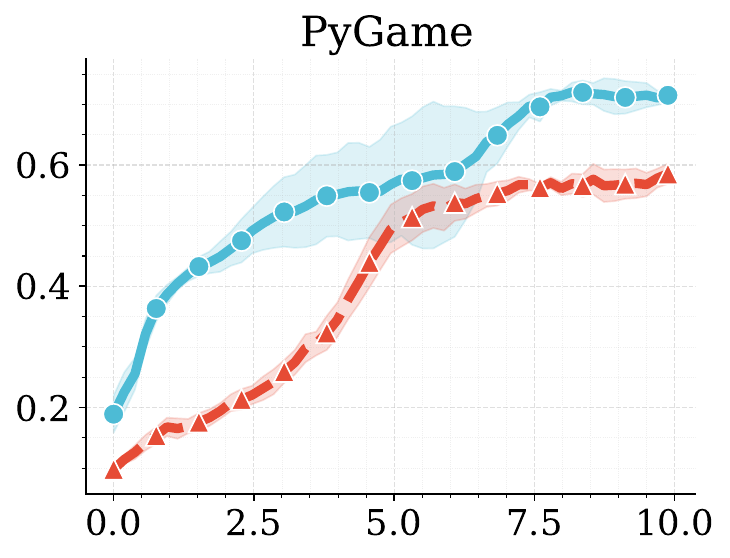}%
    \includegraphics[width=0.2\textwidth]{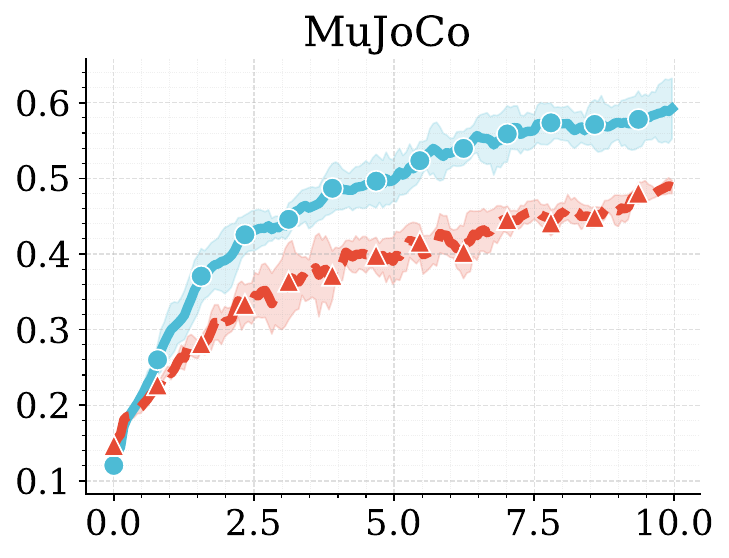}%
    \includegraphics[width=0.2\textwidth]{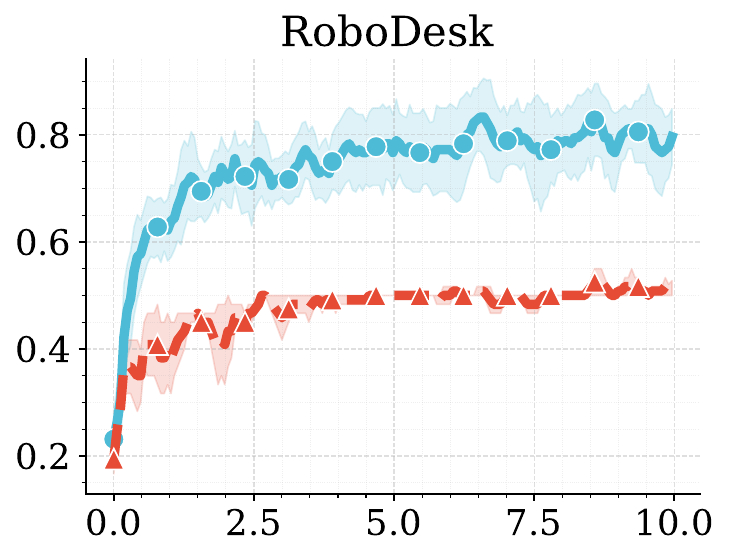}
    \vspace{-0.6cm}
    \caption{\textbf{Per-domain aggregate performance across all 10 MMBench domains.} Average normalized score of \hlSmall{\textbf{MR.Q}} (solid, \textcolor{teal}{teal}) versus \hlMedium{\textbf{\textit{Newt}}} (dashed, \textcolor[HTML]{E64B35}{red}) on state-based multitask benchmarks from MMBench~\citep{Hansen2025Newt}, spanning continuous control, manipulation, locomotion, and discrete game domains. MR.Q, a model-free agent with model-based representation learning, consistently matches or surpasses the model-based \textit{Newt} baseline in sample efficiency and final performance across all domains. Shaded regions denote 95\% CIs across five seeds.} 
    \vspace{-0.3cm}
    \label{fig:learning_across_tasks}
\end{figure*}

\paragraph{Learning Across Tasks.}
We consider a multitask setting where a single agent is trained jointly across a diverse set of environments that share observation and action spaces but differ in dynamics and reward functions, following prior work on multitask RL~\citep{Hansen2025Newt}. Training is performed by interleaving experience from multiple tasks under a shared set of parameters. This setup enables knowledge sharing across tasks, but introduces several challenges: \textit{(i) non-stationarity}, as the data distribution shifts with the task mixture; \textit{(ii) interference}, as shared representations must support multiple, potentially conflicting objectives; and \textit{(iii) optimization difficulty}, as gradients from different tasks may not align. These challenges make representation learning a central bottleneck for scaling RL systems. These challenges are particularly relevant for evaluating predictive representation learning. If predictive objectives improve latent structure, temporal consistency, and feature reuse across tasks, they may alleviate optimization instability and reduce interference even in the absence of explicit planning. \autoref{fig:learning_across_tasks} shows that MR.Q consistently improves both sample efficiency and final performance across diverse multitask domains, suggesting that predictive representation learning alone can substantially improve cross-task generalization and optimization stability. Additional per-task learning curves are provided in \autoref{appndx:per_game_learning_curves}, and detailed descriptions of the multitask suites and training protocol are given in \autoref{appndx:tasks_description}. Unless otherwise specified, results are averaged over 5 seeds.

\paragraph{Training for Longer.}
\begin{wrapfigure}{r}{0.5\textwidth}
    \centering
    \vspace{-0.5cm}\includegraphics[width=0.25\textwidth]{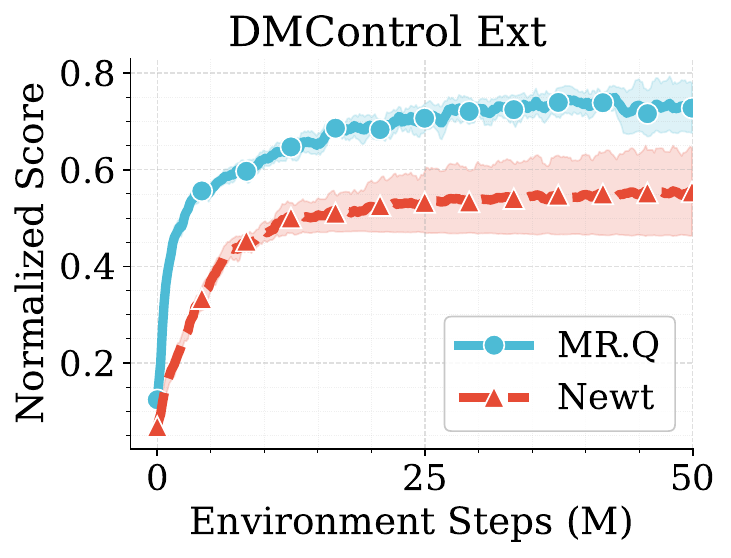}%
    \includegraphics[width=0.25\textwidth]{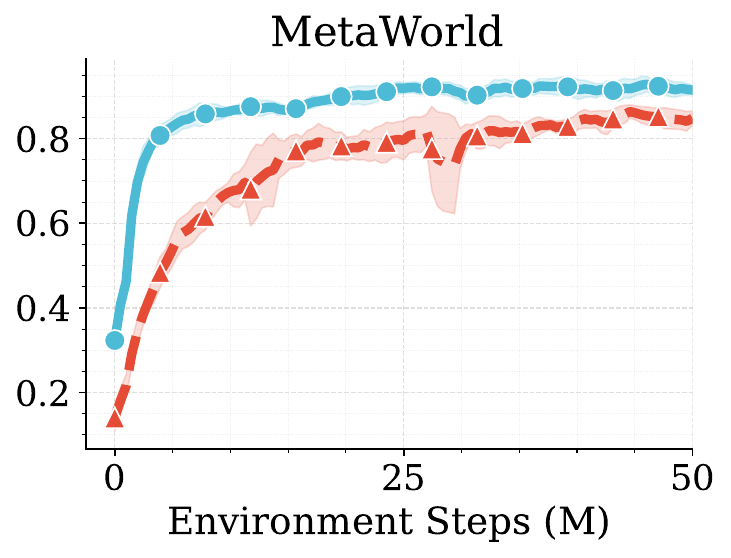}
    \vspace{-0.7cm}
    \caption{\textbf{Extended training performance (up to 50M environment steps).} 
    \hlSmall{MR.Q} sustains strong performance at scale, surpassing \hlMedium{\textit{Newt}}, indicating that gains from structured representations persist beyond the low-data regime.}
    \vspace{-0.1cm}
    \label{fig:training_longer_main}
\end{wrapfigure}
While our primary evaluation focuses on the low-data regime, it is important to assess whether the observed gains persist at larger interaction budgets. To this end, we evaluate MR.Q in extended training settings, scaling up to 50M environment steps. This allows us to analyze asymptotic performance and determine whether improvements in representation learning continue to provide benefits beyond the initial learning phase. \autoref{fig:training_longer_main} shows that MR.Q maintains strong performance as the number of interactions increases, matching or surpassing model-based approaches such as \textit{Newt} in multitask settings. This suggests that the advantages of structured representation learning are not limited to sample efficiency, but also translate to improved scalability; reinforcing the view that a model-free method equipped with structured representations provides a scalable alternative to model-based approaches, achieving strong performance across both low- and high-data regimes. 

\paragraph{Visual Observations.}

\begin{figure*}[!h]
    \centering
    \includegraphics[width=0.2\textwidth]{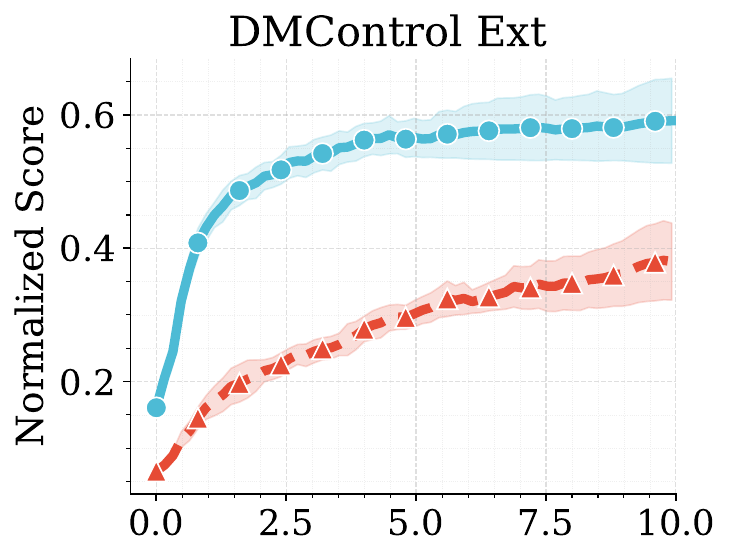}%
    \includegraphics[width=0.2\textwidth]{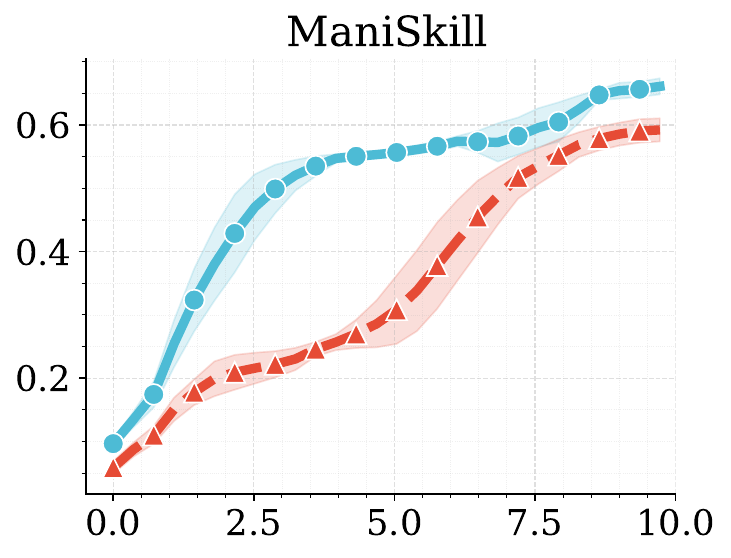}%
    \includegraphics[width=0.2\textwidth]{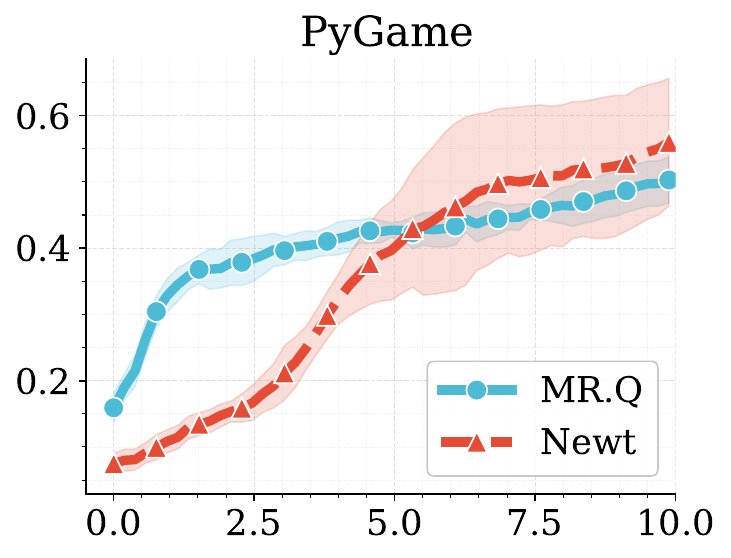}%
    \includegraphics[width=0.2\textwidth]{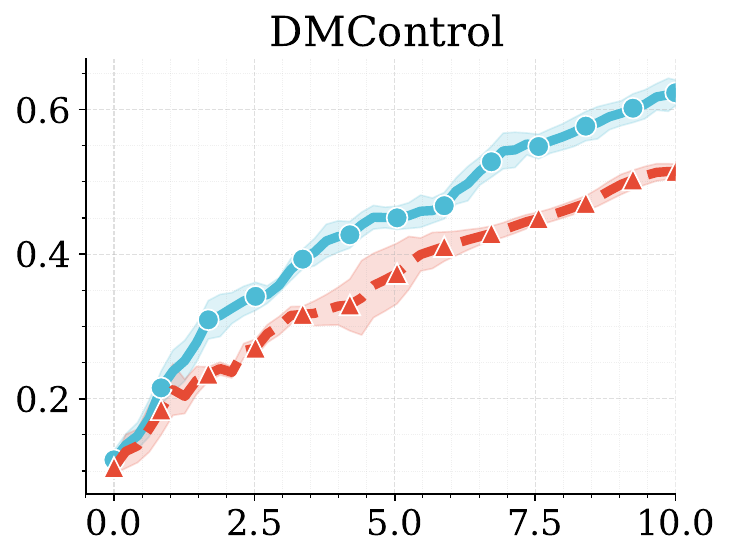}%
    \includegraphics[width=0.2\textwidth]{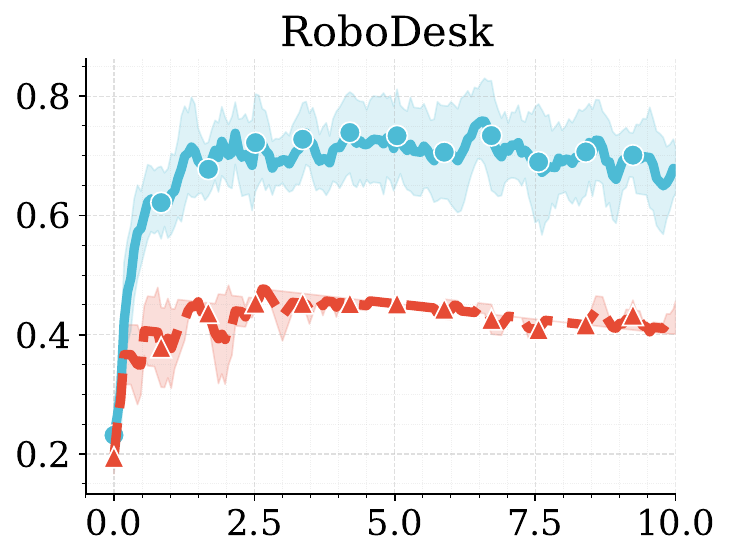}%
    \vspace{-0.31cm}
    \caption{\textbf{Pixel-based multitask learning curves across five domains.} Average normalized score of \hlSmall{\textbf{MR.Q}(solid)} and \hlMedium{\textbf{\textit{Newt}} (dashed)} using visual observations with a frozen DINOv2 encoder. MR.Q consistently achieves higher sample efficiency and final performance, demonstrating that its predictive auxiliary objectives yield better task-relevant representations in the high-dimensional input regime. Shaded regions denote 95\% CIs across five seeds. 
    } 
    \vspace{-0.3cm}
    \label{fig:vision}
\end{figure*}

While most experiments are conducted in the state-based setting, we additionally evaluate performance under high-dimensional visual inputs, following prior multitask benchmarks~\citep{Hansen2025Newt}. We use a pretrained DINOv2 encoder~\citep{oquab2024dinov} to extract features from pixels, which are then used by the policy and value networks. This setup removes the burden of learning representations from scratch, allowing us to isolate the role of downstream representation adaptation. Despite strong pretrained features, representation learning remains a key bottleneck in the multitask regime. The agent must adapt shared embeddings across diverse tasks, introducing interference and instability in value learning. As shown in \autoref{fig:vision}, MR.Q consistently outperforms \textit{Newt} across all domains, achieving higher sample efficiency and final performance. These results highlight that the benefits of structured representation learning extend beyond low-dimensional settings. Even with powerful pretrained encoders, predictive objectives remain important for learning representations that support effective multitask learning under high-dimensional inputs.

\subsection{Analyses}

To rigorously isolate the mechanisms driving performance and assess the structural integrity of the learned representations, we compare MR.Q against an encoder-free baseline (TD3) to isolate the impact of model-based representation learning. In this ablation, the encoder is removed entirely, and the actor receives a direct concatenation of the raw low-dimensional state and the 512-dimensional language instruction embedding as input, while the critic additionally receives the raw action. 

\begin{figure*}[!h]
    \centering
    \includegraphics[width=1.0\textwidth]{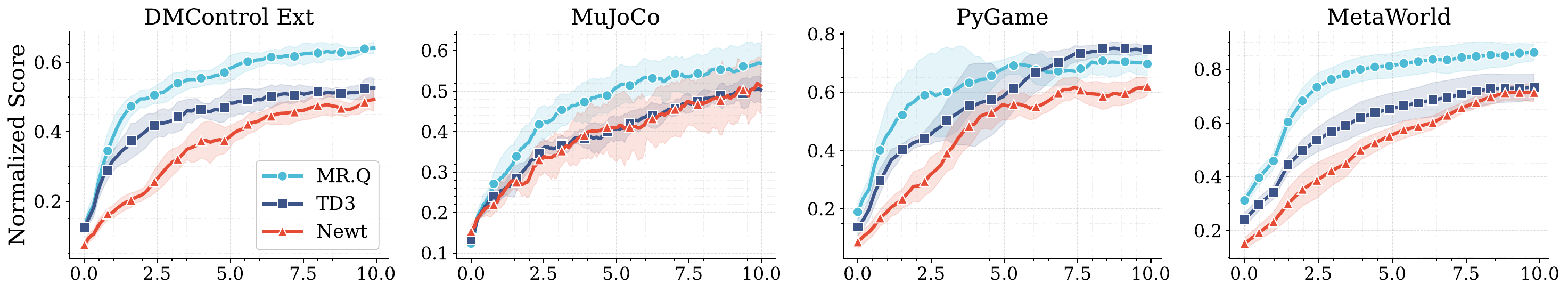}%

    \vspace{-0.32cm}
    \caption{\textbf{Performance comparison across benchmark suites.} Per-domain aggregate performance for \hlSmall{MR.Q}, the encoder-free baseline (TD3), and \hlMedium{\textit{Newt}} across four MMBench domains.}
    \vspace{-0.3cm}
\label{fig:mrq_td3_newt}
\end{figure*}

\paragraph{Performance Comparison.} We evaluate the performance of MR.Q alongside the encoder-free baseline (TD3) and \textit{Newt} as shown in \autoref{fig:mrq_td3_newt}. Our results demonstrate that MR.Q outperforms the encoder-free baseline in three out of four domains while achieving comparable results in the remaining one, demonstrating overall superior performance and sample efficiency. Interestingly, our results show that even the encoder-free baseline consistently matches or outperforms \textit{Newt} across all four domains. This indicates that a well-tuned model-free architecture utilizing raw low-dimensional states and language instruction embeddings constitutes a highly competitive baseline while offering superior computational efficiency compared to a model-based RL approach. 

\begin{figure*}[!h]
    \centering
    \includegraphics[width=0.99\textwidth]{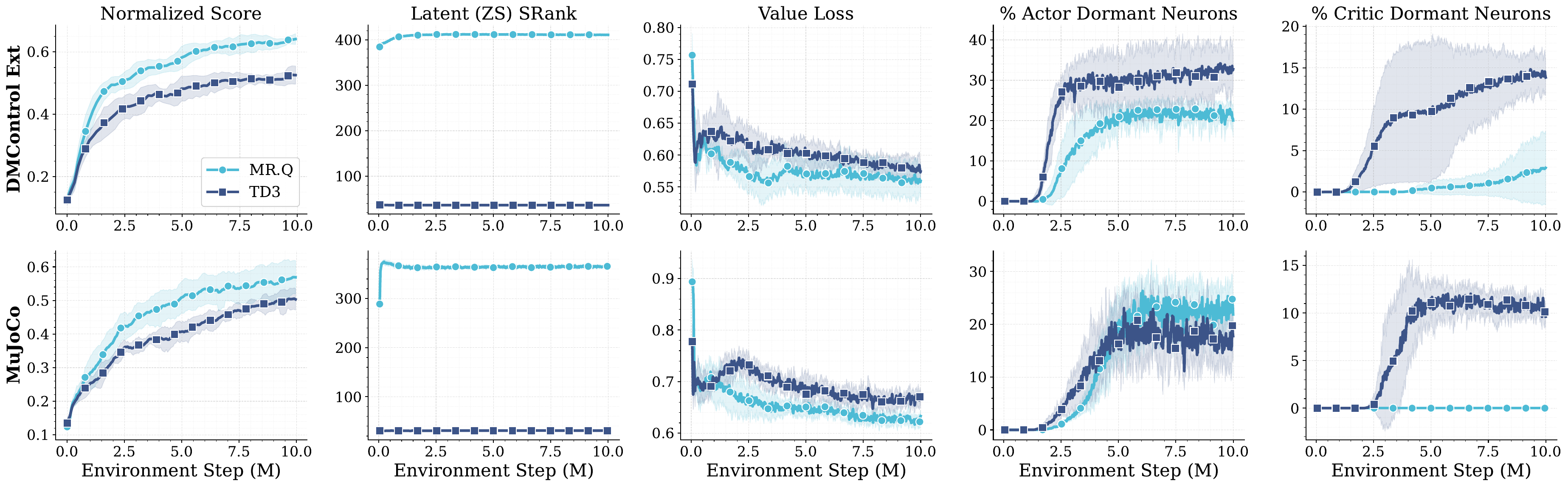}%
    \vspace{-0.35cm}
    \caption{\textbf{Empirical analyses for the effect of representation learning.} Comparison of MR.Q against an encoder-free baseline (TD3). From left to right: aggregate return across task sets, state representation SRank, value loss, and dormant neuron fractions in the actor and critic.
}
    \vspace{-0.1cm}
    \label{fig:analysis}
\end{figure*}

Notably, this finding highlights the inherent robustness of model-free RL in multitask regimes, suggesting that explicit world-modeling may not be a strict prerequisite for handling multitask RL. Beyond aggregate performance, the encoder-free baseline facilitates a diagnostic evaluation of how learned representations modulate underlying learning dynamics when compared against MR.Q. We analyze these effects across three key dimensions. These findings are summarized in \autoref{fig:analysis} and discussed in detail below. Results averaged over five seeds, shaded areas represent 95\% CIs.

\paragraph{Representation Geometry.} 

\begin{figure}[!t]
    \centering
    \includegraphics[width=0.5\linewidth]{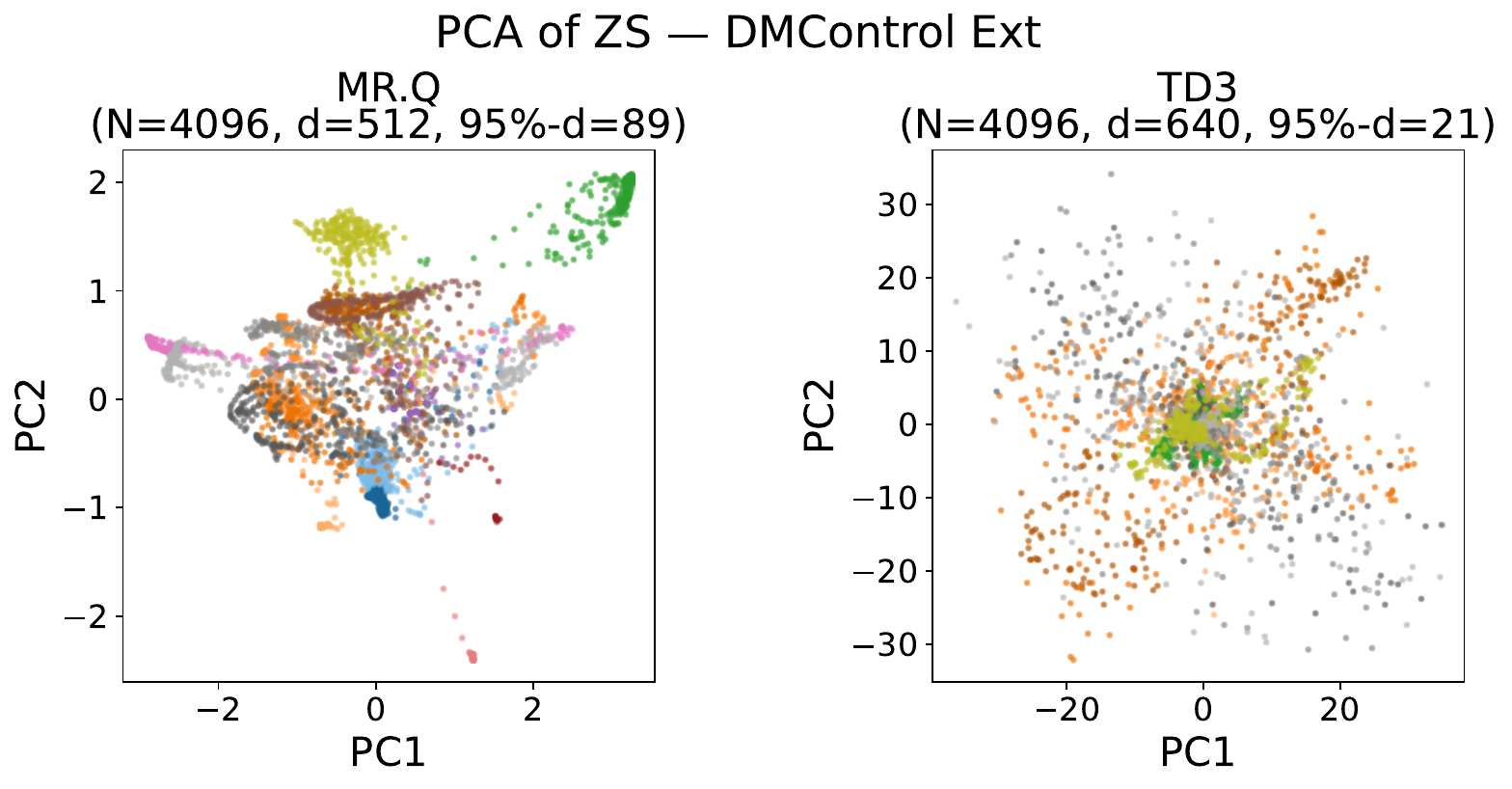}%
    \includegraphics[width=0.5\linewidth]{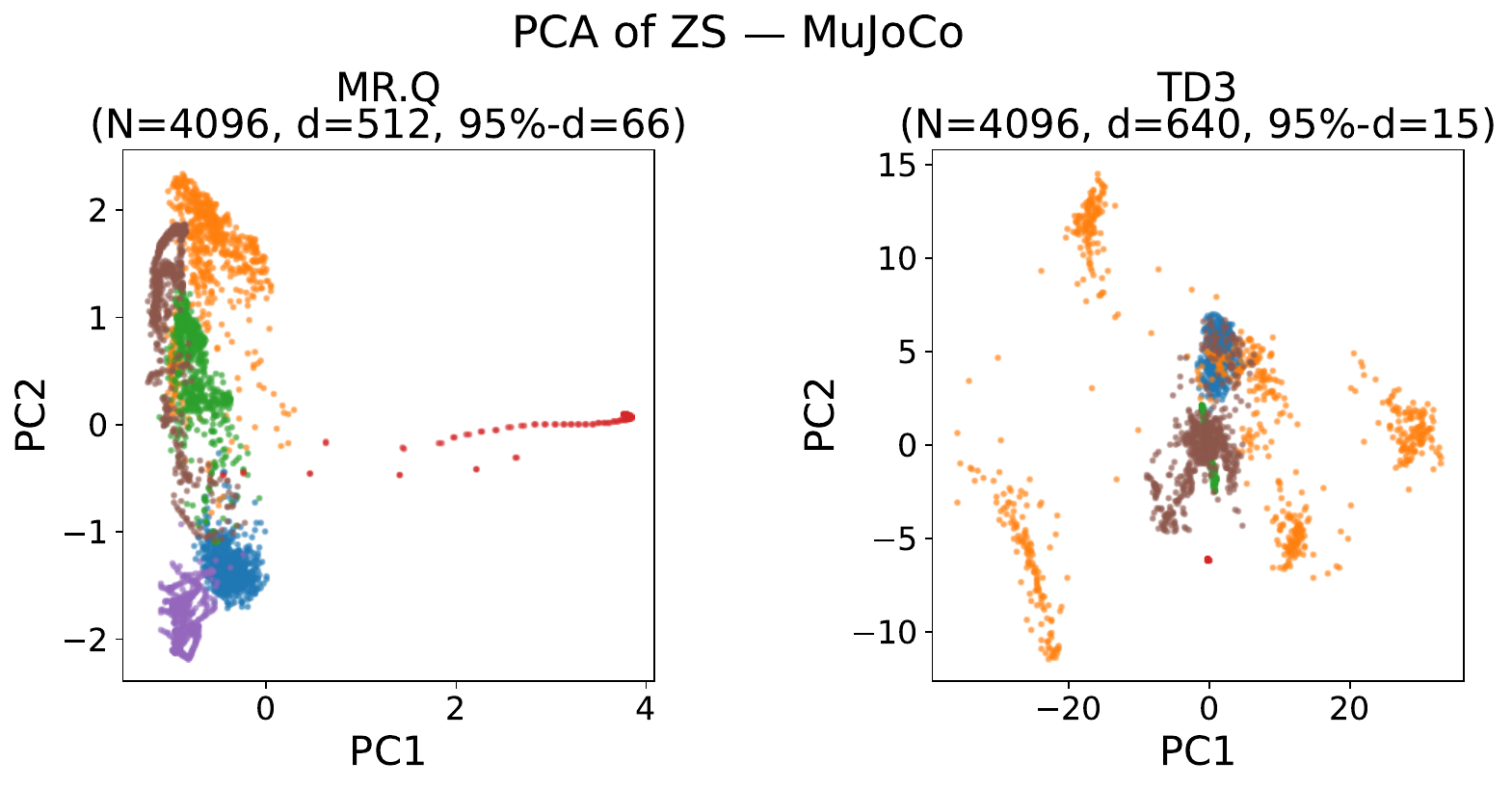}%
    \vspace{-0.35cm}
     \caption{\textbf{PCA visualization of multitask latent representations.} Two-dimensional PCA projections of latent features extracted from multitask checkpoints trained on DMControl-Ext (left) and MuJoCo (right). Each point corresponds to an observation colored by task identity. MR.Q learns structured and well-separated task representations with substantially higher effective dimensionality (95\%-d), whereas removing predictive representation learning leads to collapsed and lower-rank embeddings.
     }
    \vspace{-0.3cm}
    \label{fig:PCA_main}
\end{figure}

We evaluate feature capacity by measuring the \textbf{SRank} \citep{kumar2021implicit} of the state representations. While the encoder-free baseline TD3 uses a larger input dimensionality ($512 + d_{obs}$) than the $512$-dimensional latent space of MR.Q, its SRank is significantly lower. This suggests that raw observations result in a redundant feature space when processed without specific inductive biases. In contrast, the representation learning in MR.Q enforces a high-rank manifold that has better representational capacity. We additionally perform Principal Component Analysis (PCA) on the latent features at the end of 10M training steps to quantify the variance distribution. We measure effective dimensionality by calculating the number of principal components required to explain 95\% of the variance ($95\%$-d). Consistent with the SRank collapse, removing the encoder causes a severe representational bottleneck: across the DMControl-ext and MuJoCo suites, the $95\%$-d drops from $89$ and $66$ down to merely $21$ and $15$, respectively.
As depicted in \autoref{fig:PCA_main}, these results validate the necessity of representation learning in preserving the expressive capacity required to scale across diverse multitasks. Colors denote different tasks (12 tasks for DMControl-Ext and 6 tasks for MuJoCo); task labels are omitted in the main figure for readability, with fully annotated visualizations provided in \autoref{appndx:training_analyses}.

\paragraph{Training Dynamics}

To study how representation quality impacts optimization dynamics, we monitor the fraction of dormant neurons~\citep{sokar2023dormant,liu2025measure}, which measures the proportion of inactive units in the network. Dormant neurons indicate underutilized capacity and reduced plasticity, both of which are particularly harmful in multitask settings where agents must continually adapt to diverse and shifting objectives. MR.Q consistently exhibits a substantially lower fraction of dormant units than the encoder-free baseline, especially in the critic network where the gap becomes pronounced throughout training. In contrast, removing predictive representation learning leads to widespread critic dormancy, suggesting that the critic fails to effectively utilize the available network capacity. This degradation is accompanied by higher value losses, indicating that collapsed or poorly structured representations make value learning significantly more difficult under multitask non-stationarity.

Overall, these results suggest that predictive representation learning not only improves representation geometry, but also preserves optimization stability and network plasticity during training \citep{mayorimpact}. By maintaining expressive and active latent features, MR.Q enables the critic to make more effective use of model capacity, helping stabilize value learning across diverse multitask domains. However, competitive performance on fixed multitask benchmarks alone does not fully characterize scalability. In practice, scalable RL systems must continue to improve with increased task diversity, model capacity, data, and computation while remaining computationally efficient. In \autoref{sec:eval_sacale}, we therefore investigate how model-free RL equipped with model-based representations behaves across these scaling axes in large multitask settings.

\section{Evaluation at Scale}
\label{sec:eval_sacale}
A central question is whether model-free methods can scale as effectively as model-based approaches in multitask RL. We study this across multiple scaling axes, including task diversity, model capacity, data, update frequency, and computational efficiency.

\begin{figure*}[!h]
    \centering
    \includegraphics[width=0.2\textwidth]{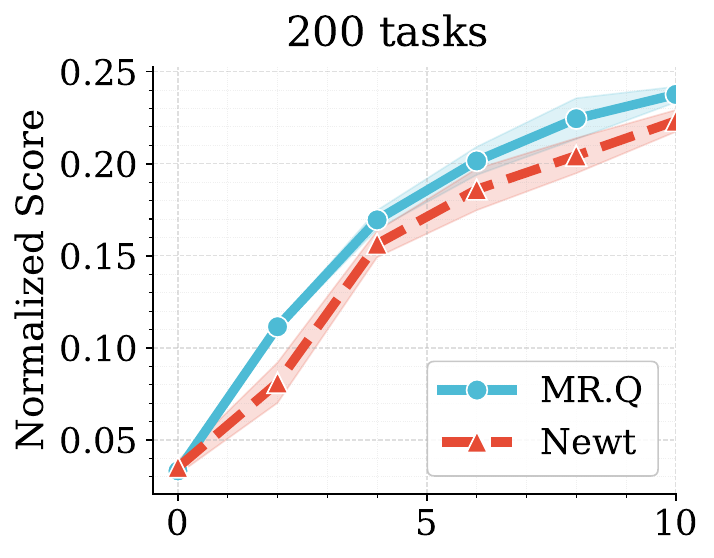}%
    \vrule width 1pt height 2.2cm
    \includegraphics[width=0.4\textwidth]{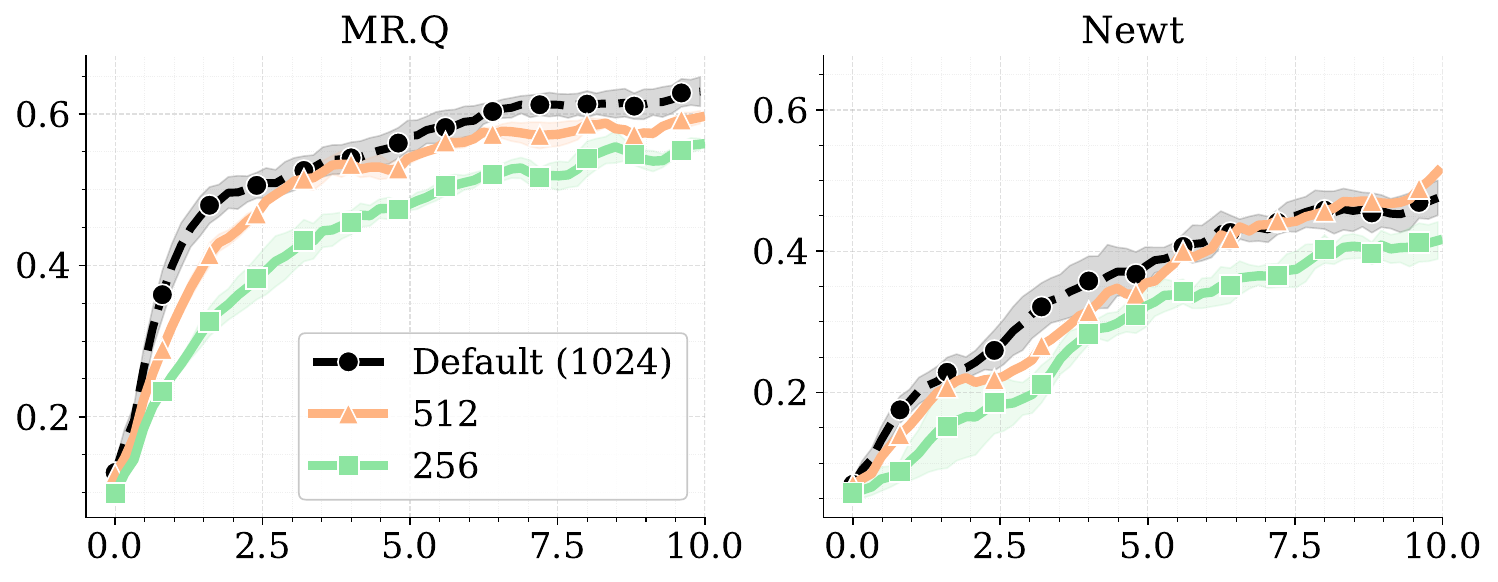}%
    \vrule width 1pt height 2.2cm
    \includegraphics[width=0.4\textwidth]{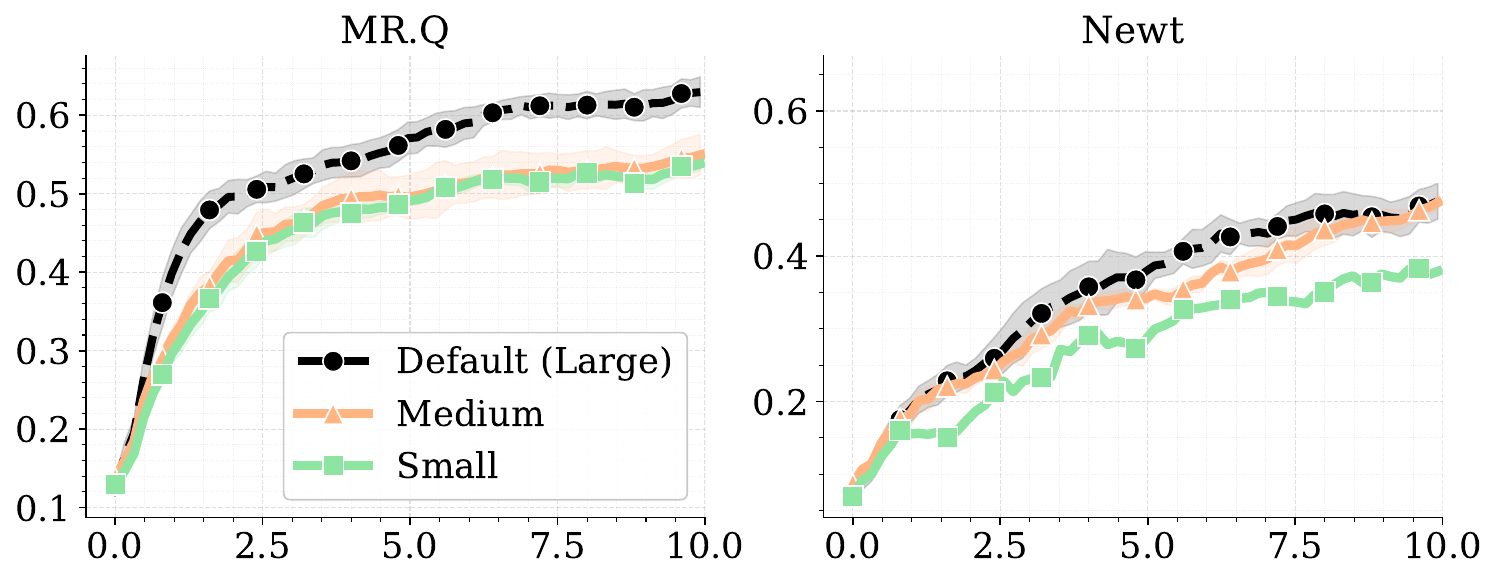}%
    \vspace{-0.2cm}
    \caption{\textit{(Left)}\textbf{ Large-scale multitask training across 200 tasks.} Normalized score throughout training on a combined benchmark of tasks spanning multiple domains. MR.Q consistently outperforms \textit{Newt} during training, while both methods converge to similar final performance. \textbf{Data and model scaling in multitask RL.} (\textit{Middle}) Data scaling: performance as a function of training data for different dataset sizes. (\textit{Right}) Model scaling: performance across model sizes. MR.Q exhibits stable scaling across both axes, while \textit{Newt} shows sensitivity to reduced data and smaller models.
    }
    \vspace{-0.3cm}
    \label{fig:ablations}
\end{figure*}

\paragraph{Towards General Multitask RL Agents.}
To evaluate scalability, we train MR.Q on a large combined benchmark of 200 tasks spanning multiple domains in a unified setting. This setting stress-tests whether structured, model-free representations can scale to the diversity required of general-purpose multitask agents, where a single model must simultaneously solve locomotion, manipulation, navigation, and arcade tasks. \autoref{fig:ablations} (left) reveals that MR.Q exhibits substantially higher \emph{sample efficiency} throughout training: at 2M environment steps it achieves a normalized score of 0.11 versus 0.08 for \textit{Newt} (+37\% relative), and maintains a consistent lead of 5--8\% across the range. This early advantage is practically significant in large-scale settings, where each additional interaction is costly. From a representation learning perspective, this suggests that model-free agents with structured latent spaces can match the representational expressiveness of world-model-based methods at scale, while requiring fewer environment interactions to do so. 

\paragraph{Model and Data Scaling.}

We analyze how multitask performance scales with data and model capacity. \autoref{fig:ablations} (\textit{middle}) shows performance as a function of available training data. Both methods improve with increased data, but exhibit different scaling behaviors. MR.Q shows consistent gains across data regimes, maintaining strong performance even with reduced data. In contrast, \textit{Newt} is more sensitive to data availability, with larger performance degradation in low-data settings. \autoref{fig:ablations} (\textit{right}) shows scaling with model size. MR.Q exhibits smooth and predictable improvements as capacity increases, indicating effective utilization of additional parameters. \textit{Newt}, however, shows weaker scaling, with smaller gains and higher sensitivity to model size. These results suggest that scaling performance is determined not only by access to more data or larger models, but also by how effectively additional capacity is utilized. MR.Q exhibits more stable scaling behavior across both axes, allowing it to better leverage increased data and model capacity.

\begin{wrapfigure}{r}{0.35\textwidth}
    \centering
    \vspace{-0.5cm}\includegraphics[width=0.341\textwidth]{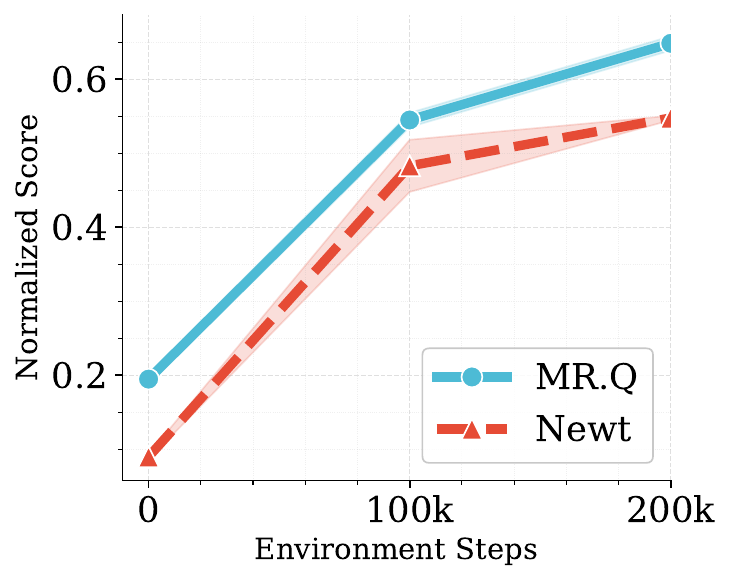}%
    \vspace{-0.3cm}
 \caption{\textbf{Few-shot finetuning on held-out tasks.} Average normalized score across 28 unseen tasks during finetuning steps from a 10M-step multitask checkpoint. \hlSmall{MR.Q} achieves 50\% higher zero-shot performance and $\sim$13\% advantage throughout training.}
    \vspace{-0.4cm}
\label{fig:finetunig}
\end{wrapfigure}

\paragraph{Scaling with Update-to-Data Ratio.}

We analyze how performance scales as a function of the update-to-data (UTD) ratio, which controls the number of gradient updates performed per environment interaction. Increasing UTD effectively increases the amount of computation applied to a fixed dataset, probing how efficiently a method can extract information from available data. \autoref{fig:utd} shows performance as a function of environment steps for different UTD values. As shown in \autoref{fig:utd} (\textit{left}), MR.Q benefits consistently from increasing UTD, with higher update regimes leading to improved performance across training. This indicates that the agent can effectively use additional gradient updates without destabilizing learning. In contrast, \textit{Newt} (\autoref{fig:utd}, \textit{right}) exhibits weaker scaling; performance improves slowly and shows diminishing returns at higher UTD.

\paragraph{Few-shot finetuning.}

To evaluate transfer to unseen tasks, we hold out a set of 28 tasks spanning multiple domains and finetune each individually using online RL, initializing from a checkpoint trained for 10M environment steps on the remaining 200 tasks. We compare MR.Q against \textit{Newt} under an identical finetuning budget of 200k steps. MR.Q provides a substantially stronger zero-shot initialization before any finetuning as shown in \autoref{fig:finetunig}. MR.Q achieves an average normalized score of 0.13 versus 0.09 for \textit{Newt}, a 50\% relative advantage, indicating that multitask pretraining with MR.Q yields more transferable representations. This advantage is preserved throughout adaptation: at 100k steps MR.Q scores 0.55 versus 0.48 for \textit{Newt} (+12.8\%), and at 200k steps 0.62 versus 0.55 (+12.9\%). At the individual task level, MR.Q outperforms \textit{Newt} on 17 of 28 held-out tasks (61\%) at the end of finetuning. These results suggest that MR.Q's structured, model-free representations learned during multitask pretraining transfer more effectively to novel tasks, enabling both a better starting point and faster convergence during adaptation.

\begin{figure*}[!h]
    \centering
    \includegraphics[width=0.2\textwidth]{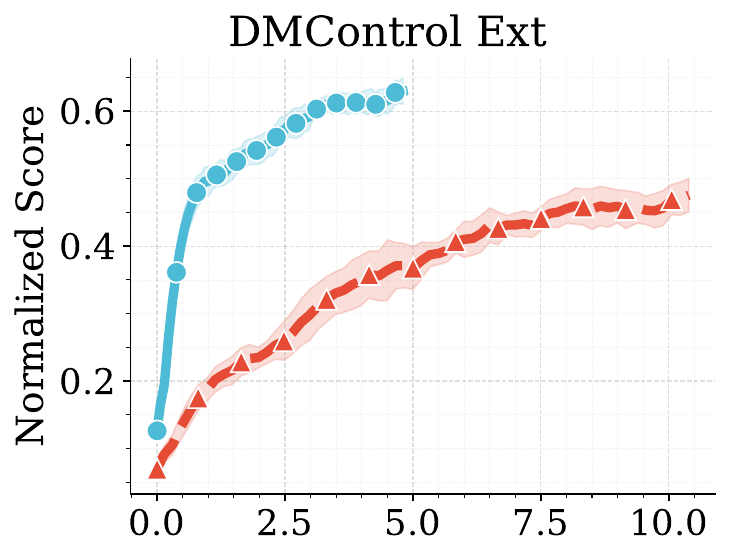}%
    \includegraphics[width=0.2\textwidth]{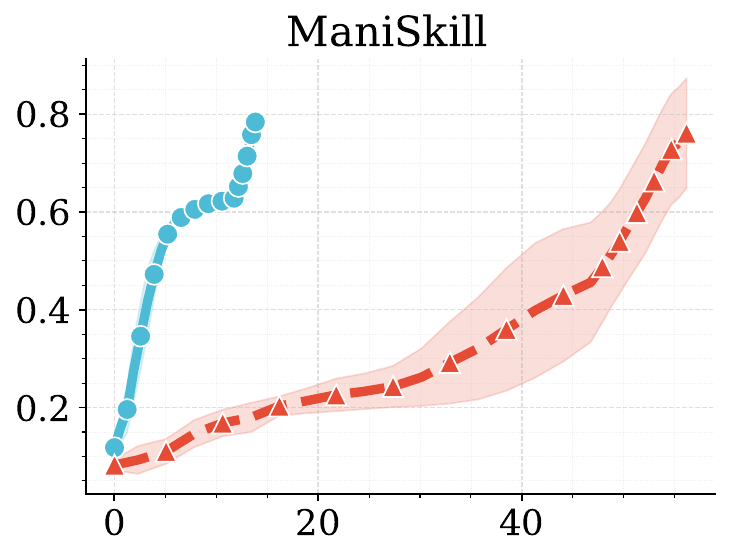}%
    \includegraphics[width=0.2\textwidth]{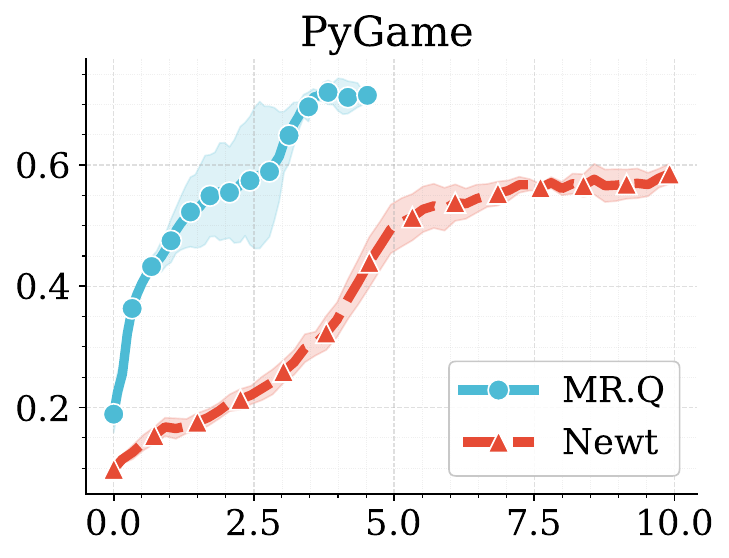}%
    \includegraphics[width=0.2\textwidth]{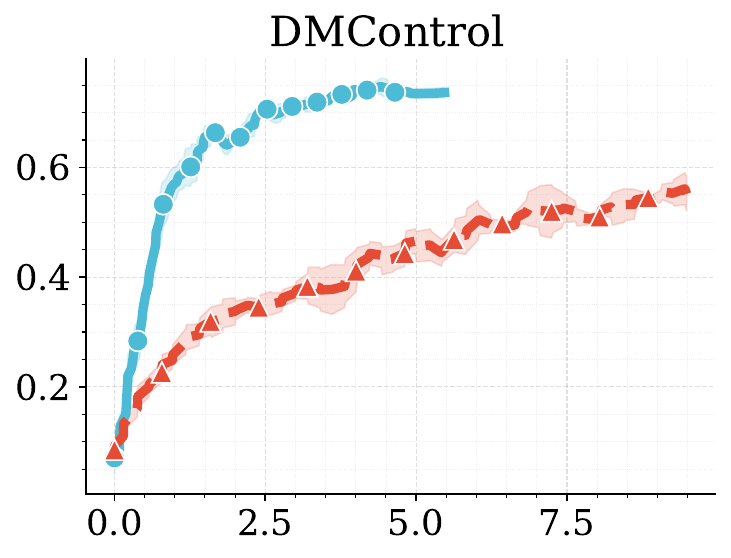}%
    \includegraphics[width=0.2\textwidth]{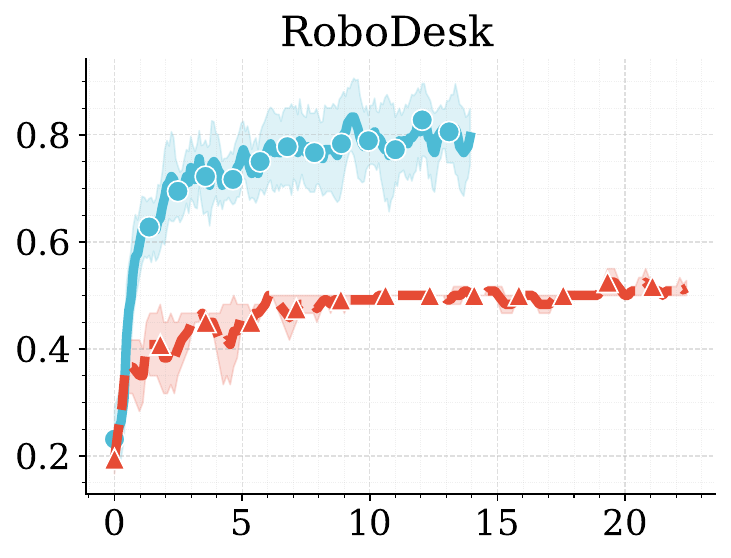}%
    \vspace{-0.25cm}
     \caption{\textbf{Wall-clock efficiency.} Normalized score as a function of wall-clock training time (hours) on five MMBench domains. \hlSmall{MR.Q} consistently reaches higher returns earlier than \hlMedium{\textit{Newt}}, a model-based baseline that incurs substantial overhead from world-model learning and latent rollout generation. Shaded regions denote 95\% CIs. All runs use a fixed budget of 10M environment steps. 
     }   
    \vspace{-0.3cm}
    \label{fig:computationa_impact}
\end{figure*}

\paragraph{Computational Impact.}
We evaluate wall-clock efficiency by measuring performance as a function of training time. While standard deep RL evaluations emphasize sample efficiency, methods with similar interaction budgets can differ substantially in time-to-performance. Model-based approaches incur additional overhead from learning dynamics models and performing latent rollouts, which slows down training despite their strong sample efficiency. In contrast, MR.Q avoids explicit planning and simulation, learning structured representations directly from data. As a result, MR.Q achieves faster improvement in performance per unit of time, reaching higher returns significantly earlier than model-based baselines, as shown in \autoref{fig:computationa_impact}. This highlights that gains in sample efficiency for world-model approaches often come at the cost of increased computational overhead. These differences have important practical implications, as higher compute requirements translate into longer training times, increased energy consumption, and reduced accessibility.

\section{Lessons and Opportunities}
\label{sec:multitask_rl}
Scaling deep RL to large and diverse multitask settings remains a central challenge. In this work, we studied the role of model-based representations in the multitask RL setting and showed that a simple model-free approach augmented with predictive objectives can match or surpass a recent large-scale world-model baseline while substantially reducing computational overhead and improving wall-clock efficiency (see \autoref{fig:computationa_impact}). Our analyses further demonstrate that predictive representation learning improves representation geometry, stabilizes optimization, and enables more effective utilization of model capacity as systems scale (\autoref{sec:eval_sacale}).

More broadly, our results highlight the importance of developing sample- and compute-efficient multitask RL algorithms that can learn effectively across diverse tasks under a realistic interaction budget. In contrast to prior work that evaluates at substantially larger scales, our primary experiments are conducted in a challenging 10M interaction regime, where efficiency and representation quality become critical bottlenecks. This is particularly important in real-world settings, where interaction data is costly. Despite this restricted budget, our approach consistently matches or surpasses large-scale world-model baselines across multiple axes, including task diversity, model scaling, transfer, and wall-clock efficiency. These findings suggest that scalable multitask RL may depend not only on larger model size or interaction budgets, but also on learning effective representations.

\paragraph{Limitations and Future Work.}
Our study focuses primarily on continuous-control multitask benchmarks, and it remains unclear how these findings extend to more diverse domains such as long-horizon environments. In addition, although our analyses suggest that predictive objectives improve representation quality and scaling behavior, the mechanisms underlying these improvements are not yet fully understood. A more principled theoretical understanding could help guide the design of future sample-efficient multitask agents. An important direction for future work is to further investigate the relationship between predictive representation learning and planning. While MR.Q demonstrates that strong multitask performance can emerge without explicit planning, hybrid approaches that combine scalable model-free representation learning with latent planning or imagination-based rollouts may provide complementary benefits \citep{chang2026surprising}. More broadly, understanding how representation learning, planning, and scaling interact in large multitask systems remains an important open problem for deep RL.

\section*{Acknowledgments}
The authors would like to thank Sami Nur Islam, Walter Mayor Toro and Gopeshh Subbaraj for valuable discussions during the preparation of this work. We would also like to give special thanks to Ghada Sokar for providing valuable feedback on an early draft of the paper.

The research was enabled in part by computational resources provided by the Digital Research
Alliance of Canada (\url{https://alliancecan.ca}) and Mila (\url{https://mila.quebec}). Pablo Samuel Castro acknowledges funding from NSERC Discovery Grant. We acknowledge funding support from Google and CIFAR AI. We would also like to thank the Python community \citep{van1995python, 4160250} for developing tools that enabled this work, including NumPy \citep{harris2020array}, Matplotlib \citep{hunter2007matplotlib}, Jupyter \citep{2016ppap}, and Pandas \citep{McKinney2013Python}.

\bibliographystyle{plainnat}
\bibliography{references}

\clearpage
\appendix

\section*{\LARGE \bfseries Appendix Contents}
\addcontentsline{toc}{section}{Appendices}

\startcontents[appendix]
\printcontents[appendix]{l}{1}{\setcounter{tocdepth}{2}}

\clearpage
\section{The Use of Large Language Models}
\label{app:llm_usage}

In this paper, LLMs were used only to polish the writing of certain paragraphs in order to improve clarity and grammar. The key ideas, theoretical analysis, method design, figures, and experimental results are entirely the result of the human authors' contributions.

\section{Impact statement}
This paper presents work whose goal is to advance the field of Machine Learning, and Reinforcement Learning in particular. There are many potential societal consequences of our work, none which we feel must be specifically highlighted here.

\section{Related Work}
\label{appndx:related_work}

\paragraph{Representation Learning and World Models in RL.}
Representation learning is a central challenge in deep RL, where learned features must support value estimation, policy optimization, generalization, and stable learning under non-stationary data distributions. A large body of work studies how auxiliary objectives, contrastive learning, reconstruction, bisimulation metrics, and predictive modeling can improve learned representations~\citep{gelada2019deepmdp,laskin2020curl,yarats2021image,yarats2022mastering,zhang2021learning}. Recent analyses further show that poor representations can lead to feature collapse, dormant neurons, reduced plasticity, and unstable value learning~\citep{kumar2021implicit,fujimoto2022should,nikishin2022primacy,sokar2023dormant,obando2025simplicial,pasand2026stable}. 

Predictive objectives are also central to modern world-model approaches. Methods such as PlaNet, Dreamer, DreamerV3, TD-MPC, and TD-MPC2 learn latent dynamics models that support imagined rollouts or latent trajectory optimization for control~\citep{hafner19a,Kaiser2020Model,Hafner2020Dream,hafner2023mastering,hansen22a,hansen2023td}. These methods demonstrate strong performance and scalability across continuous-control domains, while recent large-scale systems such as Dreamer 4 and Newt extend these principles to multitask settings~\citep{hafner2509training,Hansen2025Newt}. However, these approaches require jointly learning world models, value functions, and planning components, introducing substantial computational overhead and optimization complexity.

\paragraph{Model-Free RL with Predictive Representations.}
Several recent works suggest that predictive representation learning can improve RL even without explicit planning. Methods such as SPR~\citep{schwarzerdata}, BBF~\citep{schwarzer2023bigger}, and MR.Q~\citep{fujimototowards} augment model-free RL with auxiliary predictive objectives that encourage temporal consistency and latent structure. Similar ideas have also been explored through self-predictive representations and latent dynamics supervision~\citep{nibridging,watter2015embed}. In these approaches, predictive models are used to shape the representation rather than to generate imagined rollouts or perform trajectory optimization.

Our work builds on this line of research and studies whether predictive representation learning alone can recover many of the scalability and generalization benefits commonly associated with world-model methods. Unlike Dreamer, TD-MPC2, or Newt, our approach does not use latent planning or imagination for policy improvement. Instead, predictive objectives are used exclusively as auxiliary supervision for representation learning, allowing us to isolate the role of predictive representations from explicit model-based control.

\paragraph{Multitask Reinforcement Learning.}
Multitask RL aims to train a single agent across multiple environments while enabling transfer and representation sharing across tasks~\citep{teh2017distral,yu2020meta,sodhani2021multi}. Scaling RL to multitask settings introduces significant optimization challenges, including non-stationarity, gradient interference, negative transfer, and under-utilization of model capacity~\citep{yu2020gradient,taiga2023investigating,bai2023picor,nauman2025bigger}. These issues become increasingly severe as task diversity and model scale grow.

Recent large-scale multitask systems such as TD-MPC2 and Newt suggest that world models can scale effectively across many tasks and embodiments when trained using large shared architectures and task conditioning~\citep{hansen2023td,Hansen2025Newt}. In contrast, our work demonstrates that a simpler model-free agent equipped with predictive representations can also scale effectively across multitask domains while substantially improving computational efficiency. Our findings therefore highlight representation learning itself as a key ingredient for scalable multitask deep RL.

\section{Tasks Description}
\label{appndx:tasks_description}

For all experiments, we utilize the multitask suites introduced in MMBench~\citep{Hansen2025Newt}. This benchmark encompasses 10 distinct domains and a total of 200 diverse continuous control tasks, spanning robotic manipulation, locomotion, navigation, arcade games, and classic control. A brief overview of each domain is provided below. Full task specifications and configuration details can be found in the original MMBench benchmark~\citep{Hansen2025Newt}.
\paragraph{MuJoCo.}
The MuJoCo ~\citep{todorov2012mujoco} serves as a standard benchmark for continuous control in reinforcement learning. It comprises a variety of simulated robotic locomotion tasks, ranging from lower-dimensional kinematic problems (e.g., \textit{HalfCheetah}) to complex, high-dimensional control challenges involving severe contact dynamics (e.g. \textit{Ant}). Following MMBench, we utilize the v4 environment configurations and disable early termination conditions to ensure consistency across all evaluated task domains. 
\paragraph{DMControl and DMControl Extended.}
The DeepMind Control (DMControl) suite~\citep{tassa2018deepmind} provides a standardized set of physics-based simulation environments, with a fixed episode length of 500 and no termination conditions. DMControl Extended is an extended task set based on the original DMControl, include 11 custom tasks previously proposed by~\citet{hansen2023td}.
\paragraph{MetaWorld.}
MetaWorld~\citep{yu2020meta} is a benchmark designed for multitask and meta-reinforcement learning, focusing exclusively on simulated robotic manipulation tasks. This domain consists of 50 diverse manipulation tasks that share a unified observation and action space. Due to a known simulation issue, the \textit{Shelf Place} is excluded, yielding a final set of 49 tasks for this domain.

\paragraph{ManiSkill.}
ManiSkill3~\citep{taomaniskill3} is a comprehensive physics-based benchmark focused on complex robotic control. This domain encompasses a diverse array of tasks and robotic morphologies, spanning tabletop manipulation, quadruped locomotion, whole-body humanoid control, and mobile manipulation. Additionally, it includes reimplementations of widely adopted control environments from the MuJoCo and DMControl suites.
\paragraph{Pygame.}
Pygame consists of 22 tasks spanning 14 unique
arcade-style environment. These tasks exhibit significant heterogeneity in their core objectives, episode horizons, state-action dimensionalities, and underlying reward structures. MMBench enforce a fixed episode length across all tasks and disable early termination conditions.
\paragraph{Box2D.}
The Box2D suite~\citep{brockman2016openai} utilizes a 2D physics engine to simulate rigid body dynamics. It encompasses a well-known set of classic control, navigation, and locomotion tasks, such as \textit{LunarLander}. While the Box2D tasks were originally designed for low-dimensional state observations, MMBench modernizes the implementation by introducing support for high-dimensional visual observations.
\paragraph{RoboDesk.}
RoboDesk~\citep{kannan2021robodesk} is a specialized suite of robotic manipulation tasks designed explicitly for multitask reinforcement learning research. The benchmark features 9 distinct object manipulation tasks situated within a single, unified desk-themed environment, where all tasks share a common observation and action space.
\paragraph{Atari.}
Based on the Arcade Learning Environment (ALE)~\citep{bellemare2013arcade}, the Atari domain serves as a rigorous testbed for RL algorithms across a wide spectrum of simulated classic Atari 2600 games. More recently, \citet{farebrother2024cale} proposed a non-linear continuous-to-discrete action transformation that extends support to algorithms operating within continuous action spaces. MMBench utilizes this continuous variant of the Atari domain.

\paragraph{OGBench.}
OGBench~\citep{parkogbench} is a benchmark tailored for evaluating goal-conditioned RL and offline RL. Because it was not originally designed for standard online RL, MMBench adapts these environments by introducing redefined dense reward functions and ensuring all necessary task information is fully integrated into the observation space (\textit{e.g.} goal position).

\section{\textit{MR.Q} algorithm: Model-based Representations for Q-learning}
\label{appndx:mrq}

TD3~\citep{fujimoto2018addressing} is a model-free off-policy actor--critic algorithm for continuous control that improves stability through twin critics, delayed policy updates, and target policy smoothing. In its standard form, TD3 operates directly on environment observations without learning an explicit latent representation encoder.

\hlSmall{MR.Q}~\citep{fujimototowards} extends TD3 by introducing a learned encoder together with auxiliary predictive objectives for representation learning. Observations are first encoded into a latent representation $z_t = \phi_\xi(s_t, \tau)$ using a learned encoder $\phi_\xi$. The actor and twin critics then operate directly in latent space. In addition to standard temporal-difference learning, MR.Q trains auxiliary latent models to predict future latent representations, rewards, and termination signals from $(z_t, a_t)$. The dynamics model predicts the next latent state $\hat{z}_{t+1}$, while auxiliary heads predict rewards $\hat{r}_t$ and episode termination $\hat{d}_t$. These objectives are optimized using supervised losses and backpropagated through the shared encoder.

Importantly, the learned latent models are used exclusively for representation shaping. Unlike model-based RL methods such as Dreamer \citep{Hafner2020Dream,hafner2509training}, TD-MPC2 \citep{hansen2023td}, or Newt \citep{Hansen2025Newt}, MR.Q~\citep{fujimototowards} does not perform latent rollouts, trajectory imagination, or planning. The predictive objectives instead provide dense auxiliary supervision that encourages representations to capture temporal structure, while preserving the simplicity and efficiency of model-free RL.

\section{\textit{Newt} algorithm:}
\label{appndx:newt}

\hlMedium{\textit{Newt}}~\citep{Hansen2025Newt} builds upon TD-MPC2~\citep{hansen2023td}, a model-based RL framework that combines latent world models with trajectory optimization for control. The central idea is to learn a compact latent dynamics model that supports both value estimation and planning directly in latent space.

Given an observation $s_t$, TD-MPC2 first encodes it into a latent representation
\[
z_t = h_\theta(s_t),
\]
where $h_\theta$ is a learned encoder. A latent dynamics model then predicts future latent states conditioned on actions:
\[
\hat{z}_{t+1} = f_\theta(z_t, a_t).
\]
Additional prediction heads estimate rewards and state values:
\[
\hat{r}_t = r_\theta(z_t, a_t),
\qquad
\hat{V}_t = V_\theta(z_t).
\]

The world model is trained using supervised consistency objectives across imagined latent rollouts. TD-MPC2 optimizes a multi-step latent prediction objective of the form
\[
\mathcal{L}_{\text{model}}
=
\sum_{t,k}
\Big(
\| z_{t+k} - \hat{z}_{t+k} \|^2
+
\| r_{t+k} - \hat{r}_{t+k} \|^2
+
\| V_{t+k} - \hat{V}_{t+k} \|^2
\Big),
\]
where latent states are recursively imagined through the learned dynamics model. Unlike reconstruction-based world models, TD-MPC2 operates entirely in latent space without pixel reconstruction, improving scalability and computational efficiency.

A key difference from standard actor--critic methods is that TD-MPC2 performs explicit planning using the learned latent model. At decision time, candidate action sequences $a_{t:t+H}$ are optimized using model predictive control (MPC) by maximizing predicted future returns over imagined latent trajectories:
\[
\max_{a_{t:t+H}}
\sum_{k=0}^{H}
\gamma^k \hat{r}_{t+k}
+
\gamma^{H+1}\hat{V}_{t+H+1}.
\]
This planning procedure repeatedly rolls out trajectories inside the learned world model and selects actions according to the highest predicted return.

\textit{Newt} extends these principles to massively multitask settings by training a single language-conditioned world model jointly across hundreds of tasks and embodiments. The resulting system jointly optimizes latent dynamics learning, value estimation, reward prediction, policy learning, and trajectory optimization within a shared multitask architecture.

In contrast, MR.Q \citep{fujimototowards} uses predictive latent modeling exclusively for representation learning rather than planning. Similar to TD-MPC2, observations are encoded into latent representations and auxiliary models predict future latent states, rewards, and termination signals. However, MR.Q does not perform latent rollouts for control or trajectory optimization. The predictive objectives are instead used solely as auxiliary supervision to shape the latent representation:
\[
\mathcal{L}_{\text{MR.Q}}
=
\mathcal{L}_{\text{TD}}
+
\lambda \mathcal{L}_{\text{predictive}},
\]
where $\mathcal{L}_{\text{predictive}}$ includes latent dynamics, reward, and termination prediction losses. Policy improvement remains entirely model-free and is performed through standard actor--critic optimization rather than planning.

\section{Training Protocol}
\label{appndx:protocol}
For all experiments, we follow the multitask language-conditioned training protocol introduced in MMBench~\citep{Hansen2025Newt}. A single shared agent is trained jointly across tasks spanning multiple domains and embodiments using a unified multitask architecture. Task identity is provided through language instruction embeddings \citep{radford21a}, allowing the policy and value functions to condition behavior on the current task while sharing representations across environments. Following the official benchmark implementation, language embeddings are concatenated with state or latent features and used as additional conditioning signals throughout training.

Training is performed in an off-policy setting using replay buffers that store transitions collected across all tasks. During training, minibatches are sampled uniformly from the shared replay buffer and used to jointly optimize the actor, critics, and auxiliary predictive objectives. Unless otherwise specified, all results are averaged over five random seeds.

For visual-observation experiments, we follow prior work~\citep{Hansen2025Newt,oquab2024dinov} and use a frozen DINOv2 encoder~\citep{oquab2024dinov} to extract image representations from raw pixels. These pretrained visual features provide strong semantic representations and stabilize training in the high-dimensional input regime, allowing the downstream RL algorithm to focus on multitask adaptation and control rather than learning visual representations from scratch.

Our primary evaluation focuses on a challenging low-data regime of 10M environment interactions, substantially smaller than the budgets commonly used in prior large-scale multitask world-model systems. Additional experiments evaluate longer training horizons, model scaling, transfer, and update-to-data (UTD) scaling. Evaluation follows the normalized-score protocol introduced in MMBench, aggregating performance across tasks within each benchmark suite.

\section{Scaling with UTD}
\label{appndx:scaling_utd}

The update-to-data ratio (UTD) controls the number of gradient updates performed per environment interaction and serves as an important scaling axis for evaluating data reuse efficiency. Increasing UTD effectively increases the amount of optimization performed on a fixed dataset, testing whether a method can efficiently extract information from available experience without destabilizing learning.

\begin{figure}[!h]
    \centering
    \includegraphics[width=\linewidth]{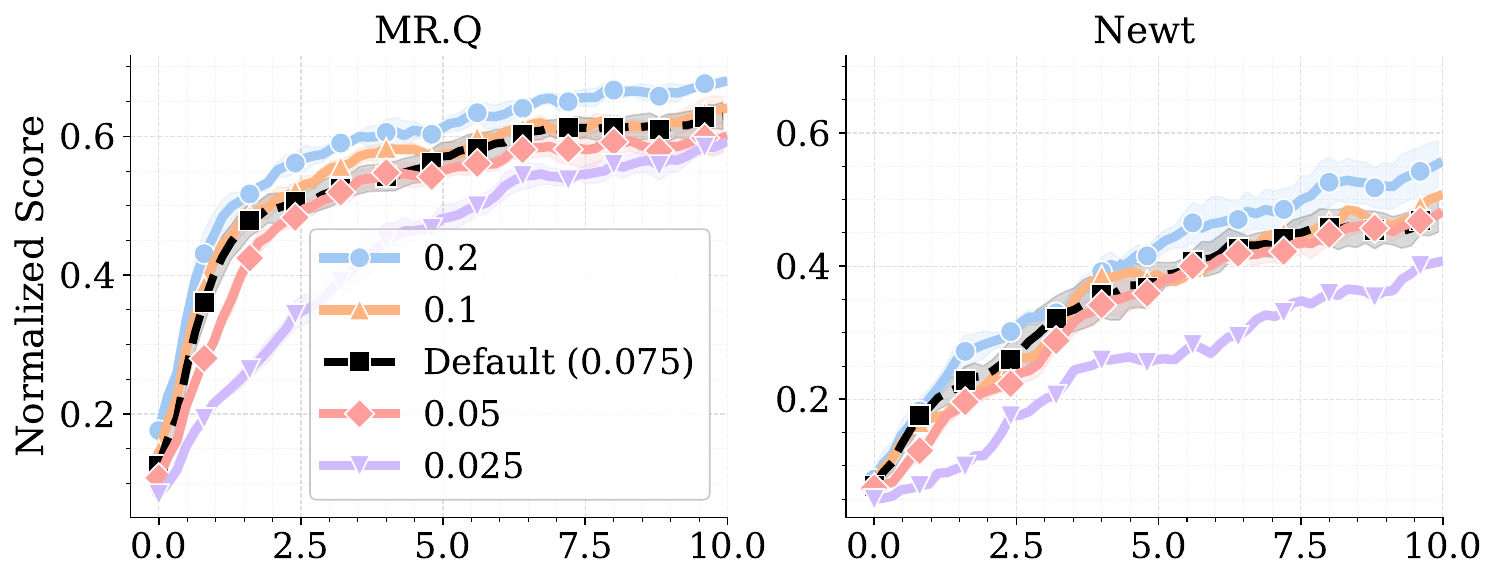}%
    \vspace{-0.4cm}
    \caption{\textbf{Scaling with UTD.} Normalized score across five multitask suites. \hlSmall{MR.Q} benefits more from higher UTD than \hlMedium{\textit{Newt}}, better data reuse.}
 \vspace{-0.2cm}
\label{fig:utd}
\end{figure}

\autoref{fig:utd} compares scaling behavior across different UTD values. MR.Q consistently benefits from larger UTD regimes, achieving improved performance as additional gradient updates are applied per interaction. In contrast, \textit{Newt} exhibits weaker gains and greater sensitivity to increased update frequency. These results suggest that predictive representation learning enables more stable and effective reuse of replay data, allowing model-free methods to better exploit additional computation under fixed interaction budgets.

\newpage

\section{PCA Analyses}
\label{appndx:training_analyses}

To further analyze the geometry of the learned multitask representations, we visualize latent features using Principal Component Analysis (PCA). We project latent representations extracted from trained checkpoints onto their top two principal components and color points according to task identity.

Across both DMControl-Ext and MuJoCo suites, MR.Q learns substantially more structured and separated latent representations than the encoder-free baseline. Predictive representation learning produces higher-rank embeddings with improved task separation and greater effective dimensionality, whereas removing representation learning leads to collapsed feature spaces with substantially reduced variance across dimensions. These results complement the quantitative analyses presented in the main paper. Together with the SRank measurements and dormant-neuron analyses, the PCA visualizations suggest that predictive auxiliary objectives improve representation diversity and preserve expressive capacity in large multitask settings.

\begin{figure}[!h]
    \centering
    \includegraphics[width=\linewidth]{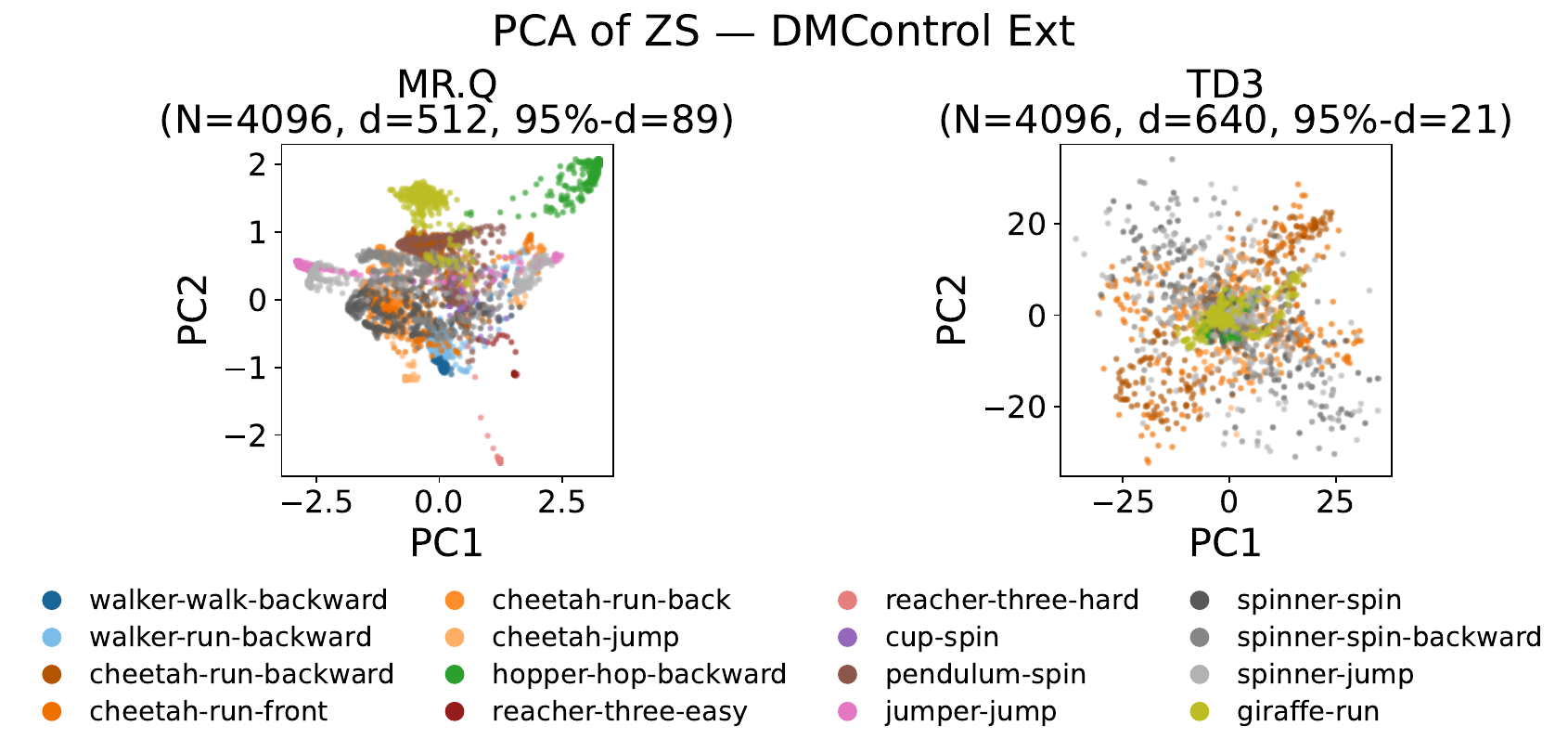}
    \vspace{-0.35cm}
     \caption{\textbf{PCA visualization on DMControl-Ext.} Two-dimensional PCA projections of multitask latent representations learned by MR.Q and the encoder-free baseline (TD3). Predictive representation learning produces substantially more structured and separated task representations.}   
    \vspace{-0.3cm}
    \label{fig:app_PCA}
\end{figure}

\begin{figure}[!h]
    \centering
    \includegraphics[width=0.85\linewidth]{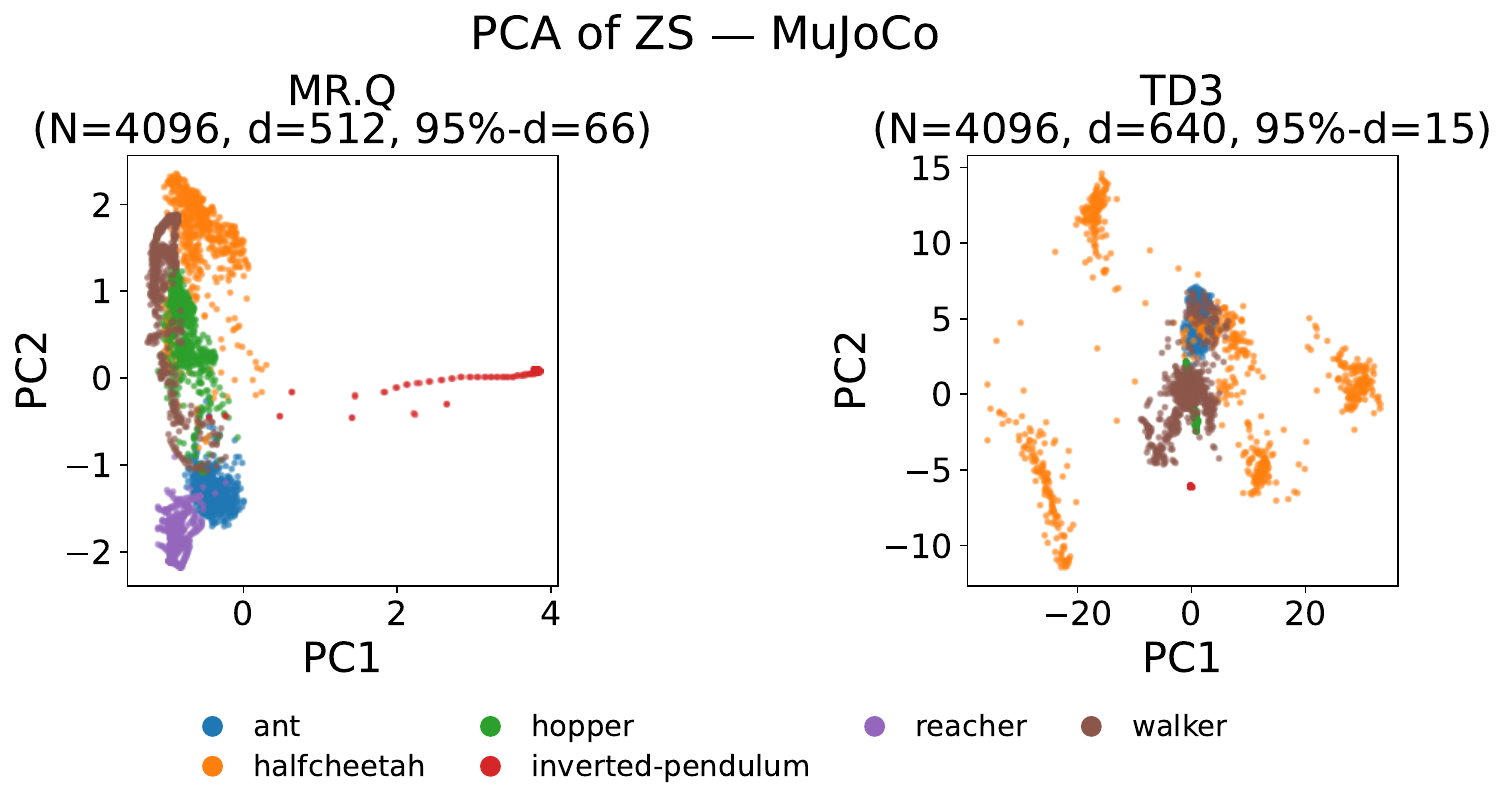}%
    \vspace{-0.35cm}
     \caption{\textbf{PCA visualization on MuJoCo.} Latent representations learned by MR.Q exhibit higher diversity and improved task separation compared to the encoder-free baseline (TD3), indicating more expressive multitask representations.}   
    \vspace{-0.3cm}
    \label{fig:app_PCA}
\end{figure}

\newpage

\section{Compute Resources}

All experiments were conducted on NVIDIA A100 GPUs using distributed Slurm-based compute clusters. Most multitask experiments were trained on a single GPU with approximately 24--48 GB of memory. Depending on the benchmark and model size, training required approximately 12--60 hours per run. Results are averaged over five seeds.

Beyond reducing training time, the computational efficiency of model-free agents equipped with predictive model-based representations has practical implications for how multitask RL systems are developed and studied. By avoiding explicit planning and latent rollout generation, our approach lowers the cost of experimentation and enables faster iteration cycles during development and finetuning. This can make large-scale multitask RL more accessible under limited compute budgets \citep{obando2020revisiting}, allowing researchers to explore architectures \citep{ceron2024mixtures,ceron2024value,sokar2025dont,liu2025neuroplastic,kooi2026hadamax}, hyperparameters \citep{andrychowicz2021what,obando2023small,ceron2024on}, and adaptation strategies without repeatedly incurring the cost of expensive model-based training pipelines.

These efficiency gains may create opportunities to scale multitask RL beyond the model sizes and experimental regimes commonly explored today. Since additional compute is not spent on planning procedures, resources can instead be allocated toward larger networks, broader task distributions, or more extensive scaling studies.

\newpage
\section{Per-tasks learning curves}
\label{appndx:per_game_learning_curves}
In addition to the aggregate results presented in the main paper, we provide per-task learning curves for all benchmark suites. These plots offer a more fine-grained view of training dynamics across individual environments.

\begin{figure*}[!h]
    \centering
    \includegraphics[width=\textwidth]{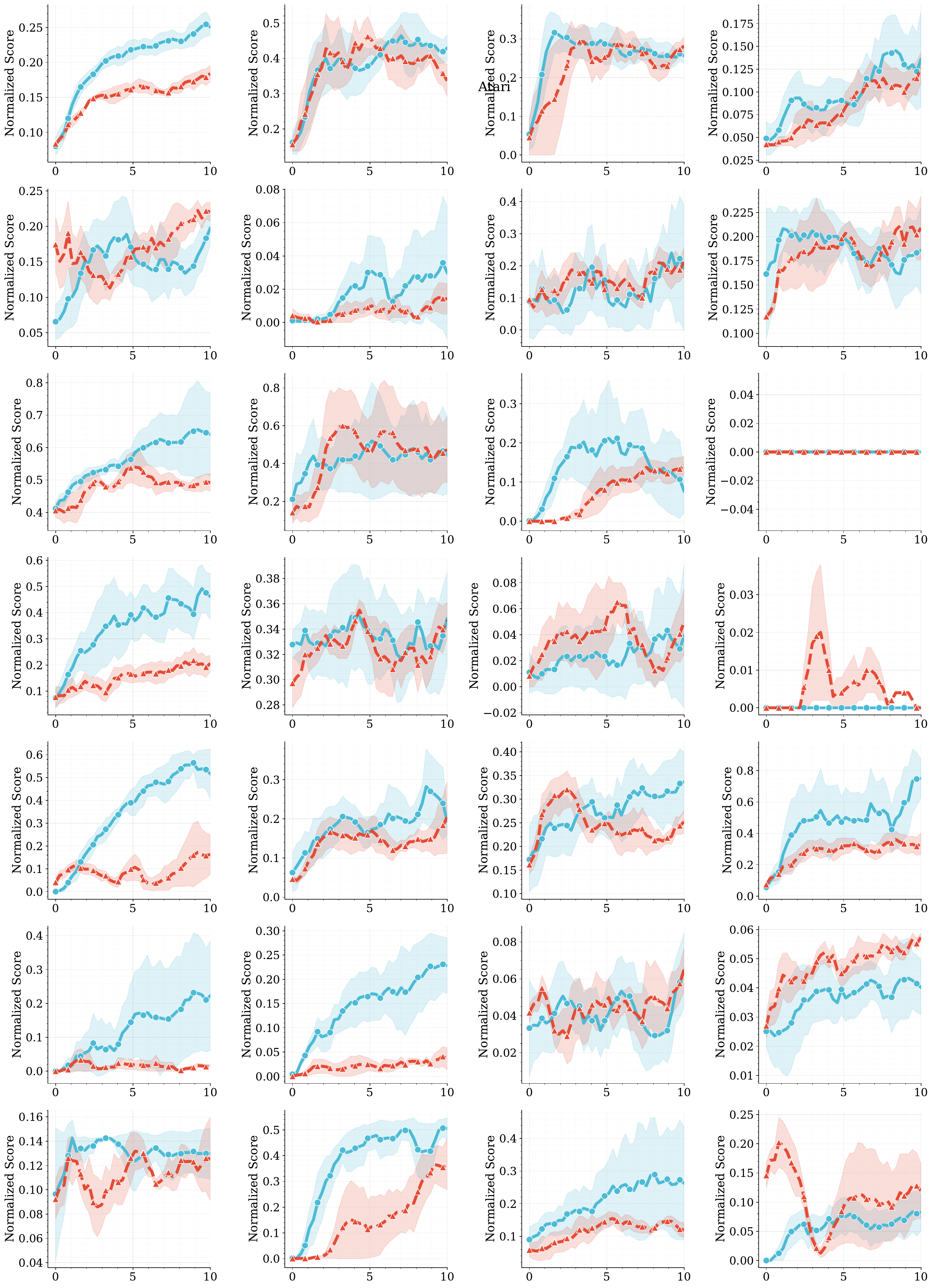}%
    \vspace{-0.35cm}
    \caption{\textbf{Atari per-game learning performance.} \hlSmall{MR.Q}, a model-free agent augmented with predictive model-based representations, consistently matches or surpasses the world-model-based approach \hlMedium{\textit{Newt}} across Atari tasks. Shaded regions denote 95\% confidence intervals (CIs).}
    \vspace{-0.3cm}
    \label{fig:atari_per_games}
\end{figure*}

\begin{figure*}[!h]
    \centering
    \includegraphics[width=\textwidth]{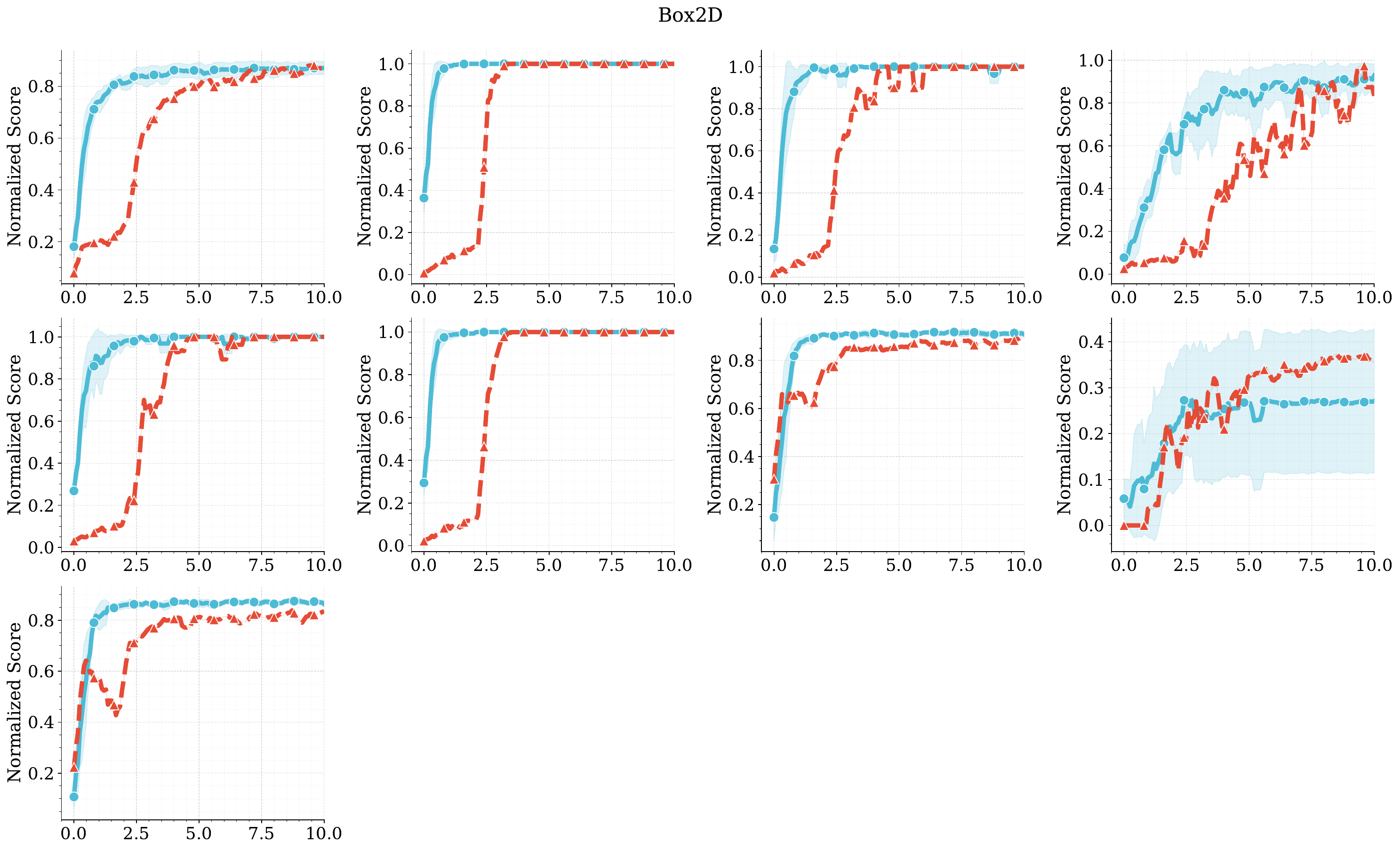}%
    \vspace{-0.35cm}
    \caption{\textbf{Box2D per-game learning performance.} \hlSmall{MR.Q}, a model-free agent augmented with predictive model-based representations, consistently matches or surpasses the world-model-based approach \hlMedium{\textit{Newt}} across Box2D tasks. Shaded regions denote 95\% confidence intervals (CIs).}
    \vspace{-0.3cm}
    \label{fig:box2d_per_games}
\end{figure*}

\begin{figure*}[!h]
\centering
\includegraphics[width=\textwidth]{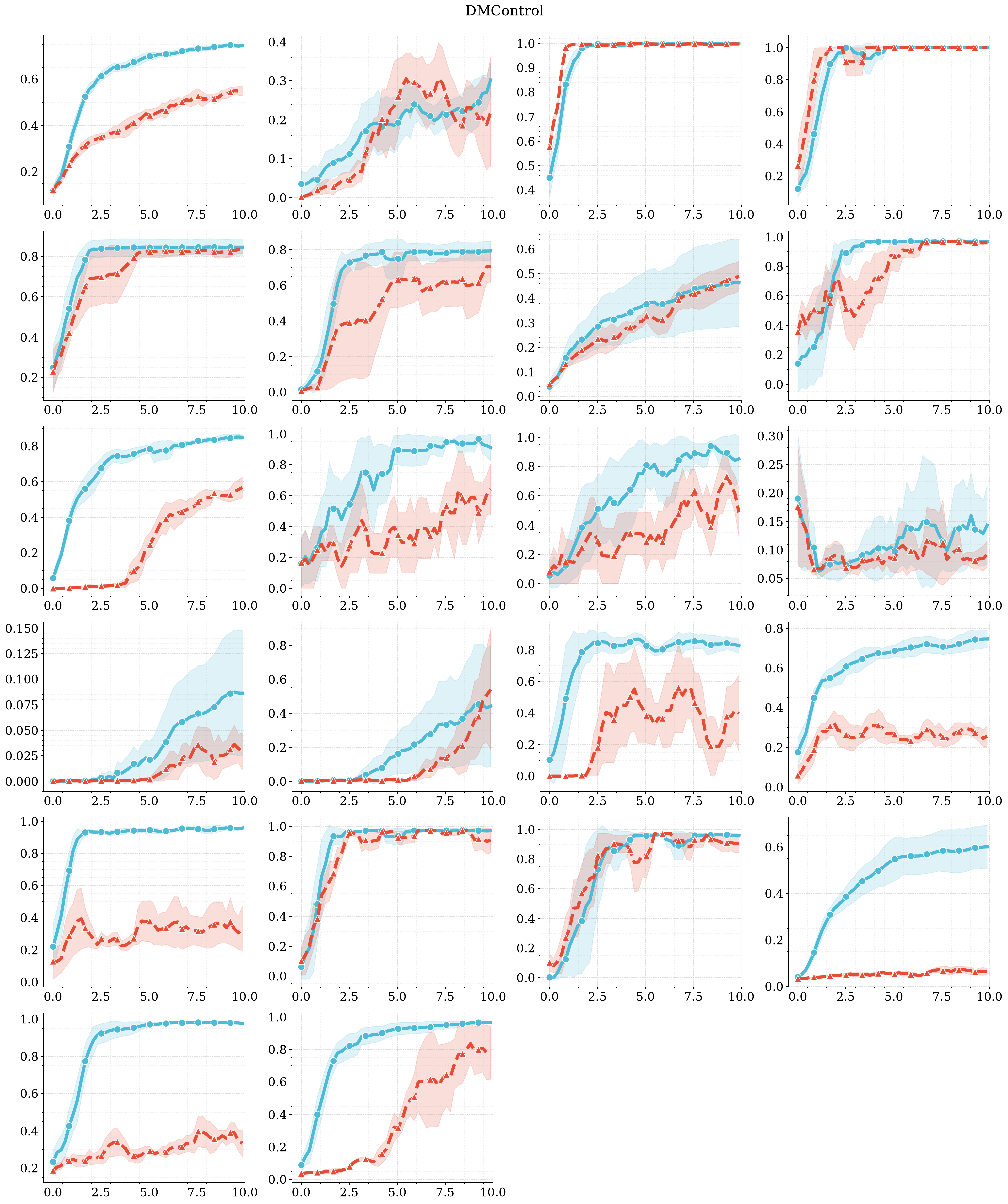}%
\vspace{-0.35cm}
\caption{\textbf{DMControl per-game learning performance.} \hlSmall{MR.Q}, a model-free agent augmented with predictive model-based representations, consistently matches or surpasses the world-model-based approach \hlMedium{\textit{Newt}} across DMControl tasks. Shaded regions denote 95\% confidence intervals (CIs).}
\vspace{-0.3cm}
\label{fig:dmcontrol_per_games}
\end{figure*}

\begin{figure*}[!h]
\centering
\includegraphics[width=\textwidth]{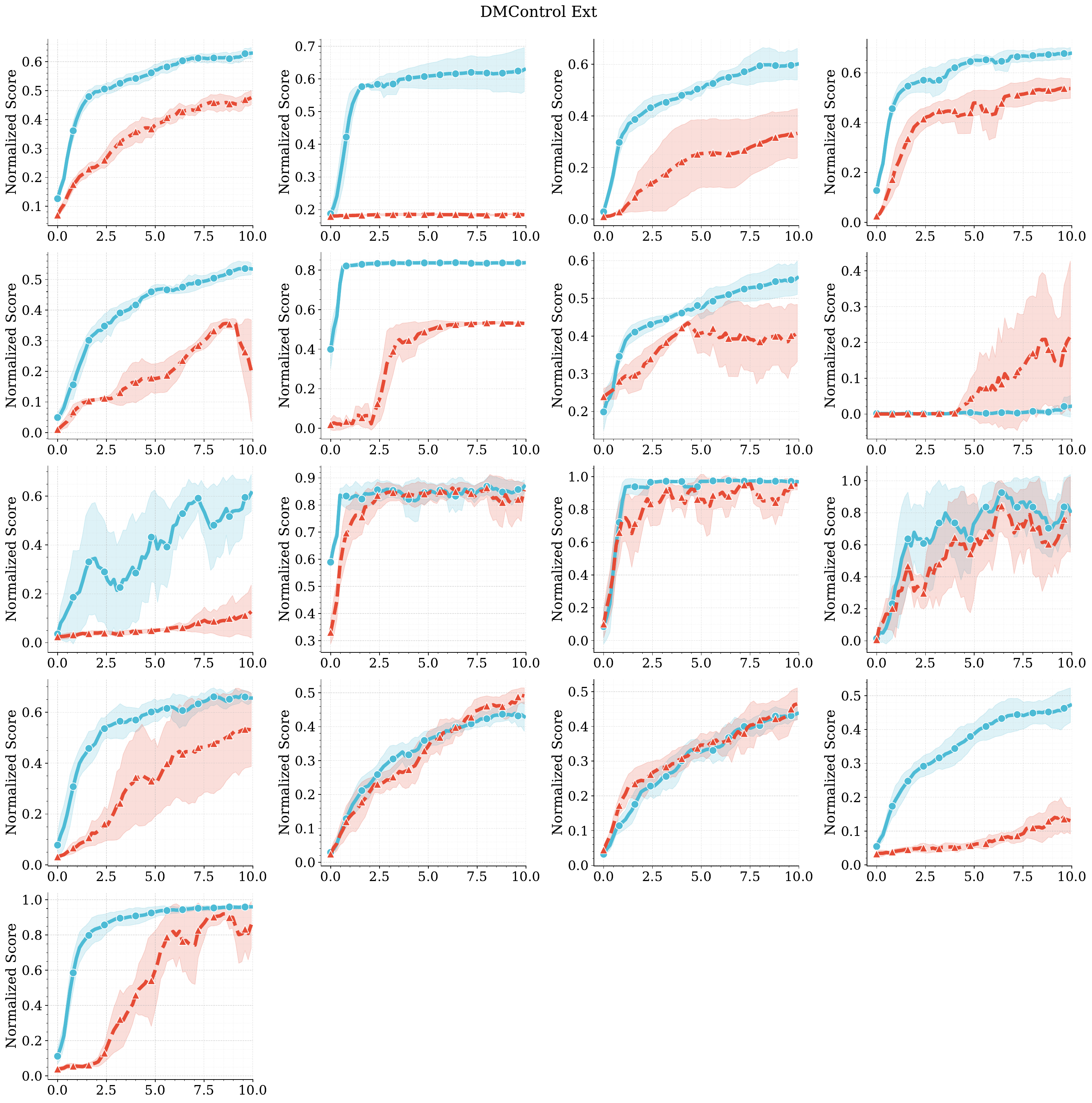}%
\vspace{-0.35cm}
\caption{\textbf{DMControl-Ext per-game learning performance.} \hlSmall{MR.Q}, a model-free agent augmented with predictive model-based representations, consistently matches or surpasses the world-model-based approach \hlMedium{\textit{Newt}} across DMControl-Ext tasks. Shaded regions denote 95\% confidence intervals (CIs).}
\vspace{-0.3cm}
\label{fig:dmcontrol_ext_per_games}
\end{figure*}

\begin{figure*}[!h]
\centering
\includegraphics[width=0.7\textwidth]{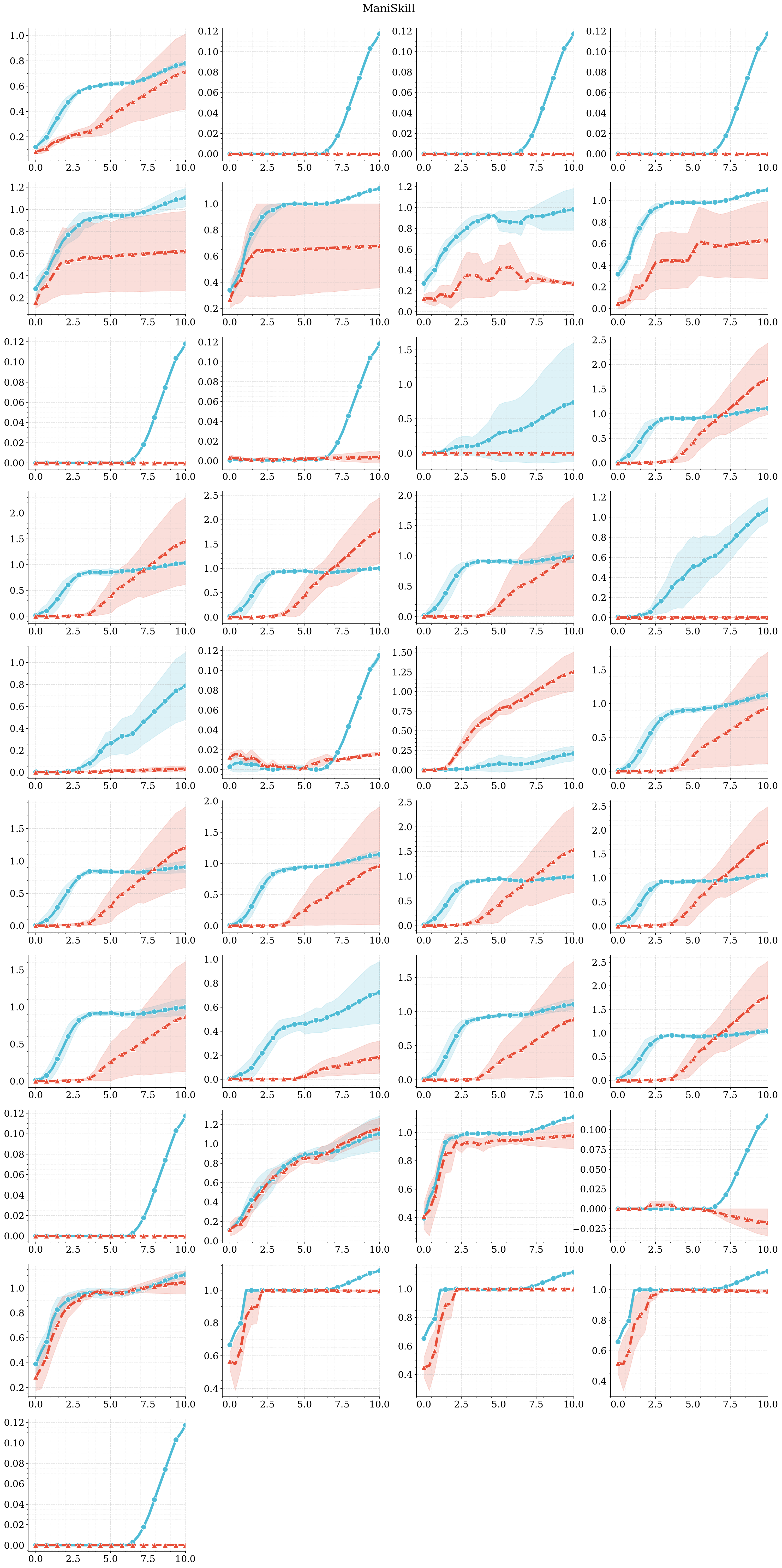}%
\vspace{-0.35cm}
\caption{\textbf{ManiSkill per-game learning performance.} \hlSmall{MR.Q}, a model-free agent augmented with predictive model-based representations, consistently matches or surpasses the world-model-based approach \hlMedium{\textit{Newt}} across ManiSkill tasks. Shaded regions denote 95\% confidence intervals (CIs).}
\vspace{-0.3cm}
\label{fig:maniskill_per_games}
\end{figure*}

\begin{figure*}[!h]
\centering
\includegraphics[width=0.6\textwidth]{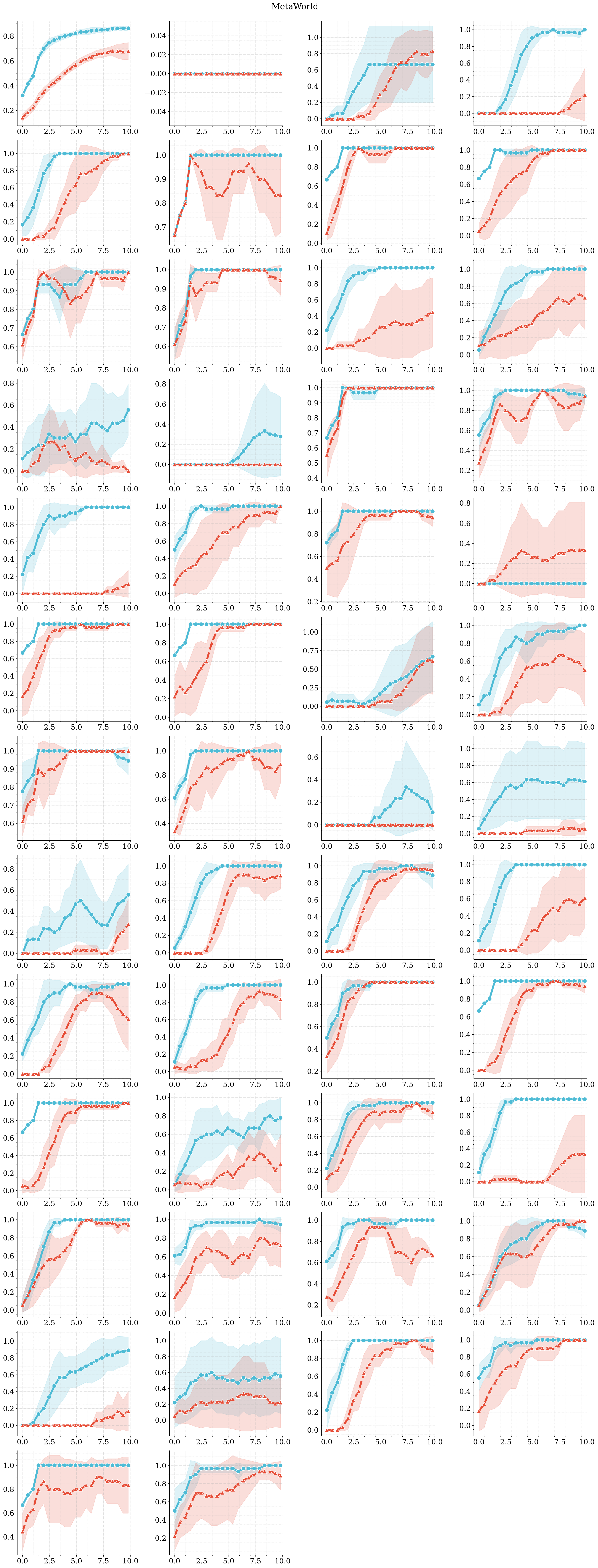}%
\vspace{-0.35cm}
\caption{\textbf{MetaWorld per-game learning performance.} \hlSmall{MR.Q}, a model-free agent augmented with predictive model-based representations, consistently matches or surpasses the world-model-based approach \hlMedium{\textit{Newt}} across MetaWorld tasks. Shaded regions denote 95\% confidence intervals (CIs).}
\vspace{-0.3cm}
\label{fig:metaworld_per_games}
\end{figure*}

\begin{figure*}[!h]
\centering
\includegraphics[width=\textwidth]{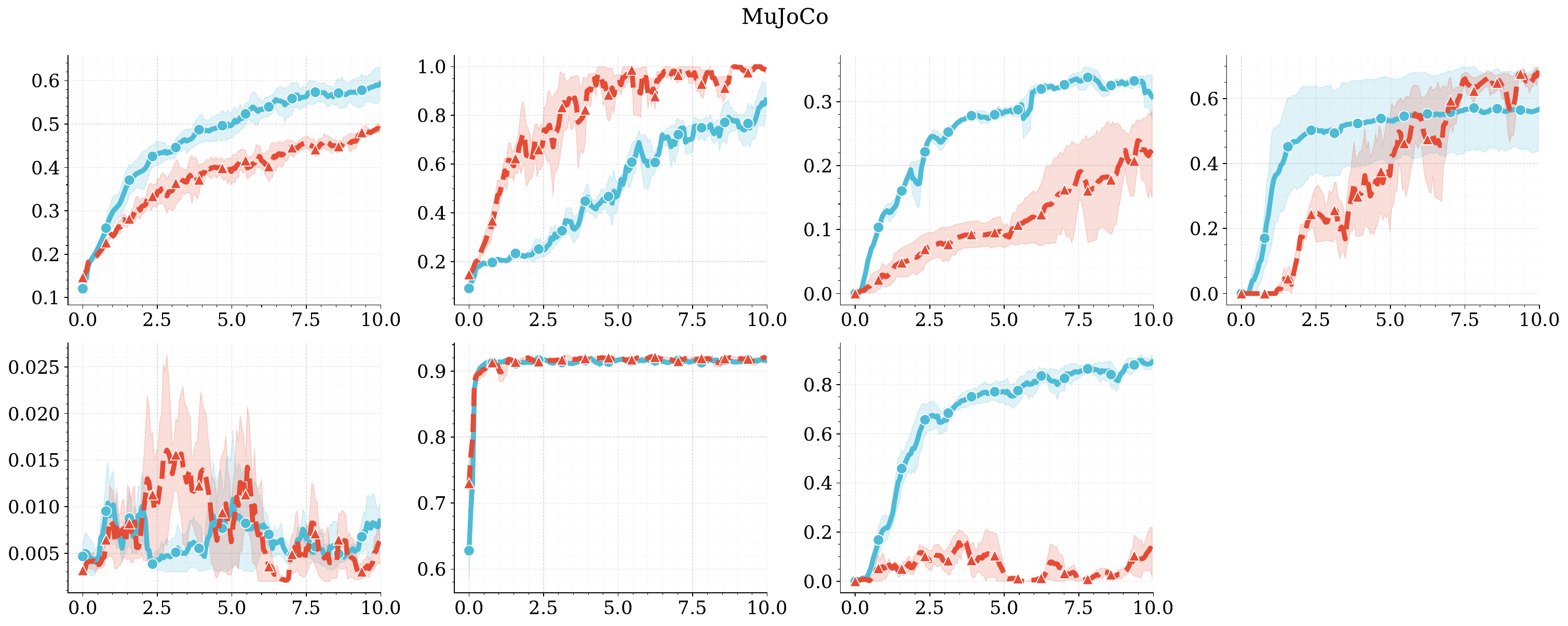}%
\vspace{-0.35cm}
\caption{\textbf{MuJoCo per-game learning performance.} \hlSmall{MR.Q}, a model-free agent augmented with predictive model-based representations, consistently matches or surpasses the world-model-based approach \hlMedium{\textit{Newt}} across MuJoCo tasks. Shaded regions denote 95\% confidence intervals (CIs).}
\vspace{-0.3cm}
\label{fig:mujoco_per_games}
\end{figure*}

\begin{figure*}[!h]
\centering
\includegraphics[width=\textwidth]{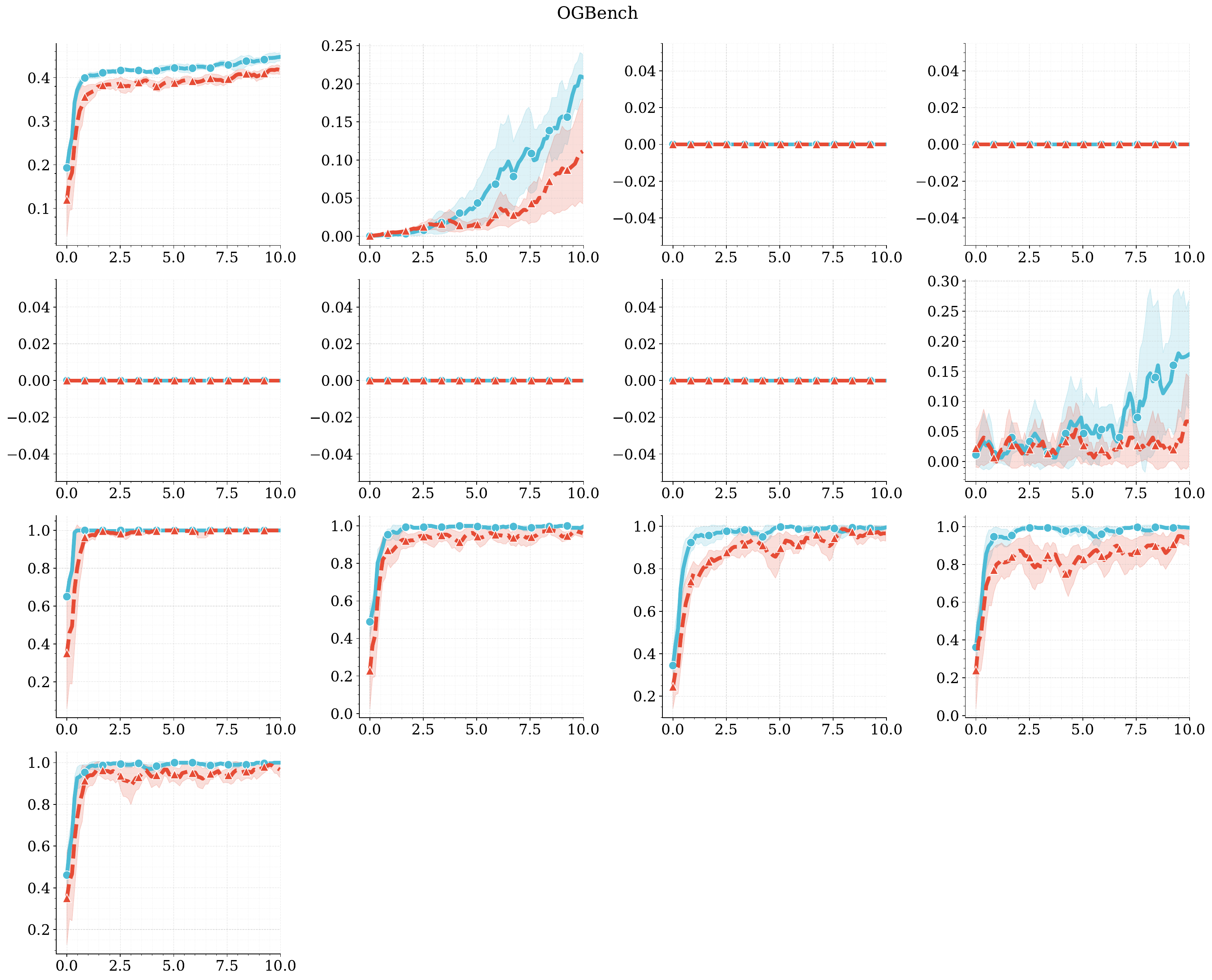}%
\vspace{-0.35cm}
\caption{\textbf{OGBench per-game learning performance.} \hlSmall{MR.Q}, a model-free agent augmented with predictive model-based representations, consistently matches or surpasses the world-model-based approach \hlMedium{\textit{Newt}} across OGBench tasks. Shaded regions denote 95\% confidence intervals (CIs).}
\vspace{-0.3cm}
\label{fig:ogbench_per_games}
\end{figure*}

\begin{figure*}[!h]
\centering
\includegraphics[width=\textwidth]{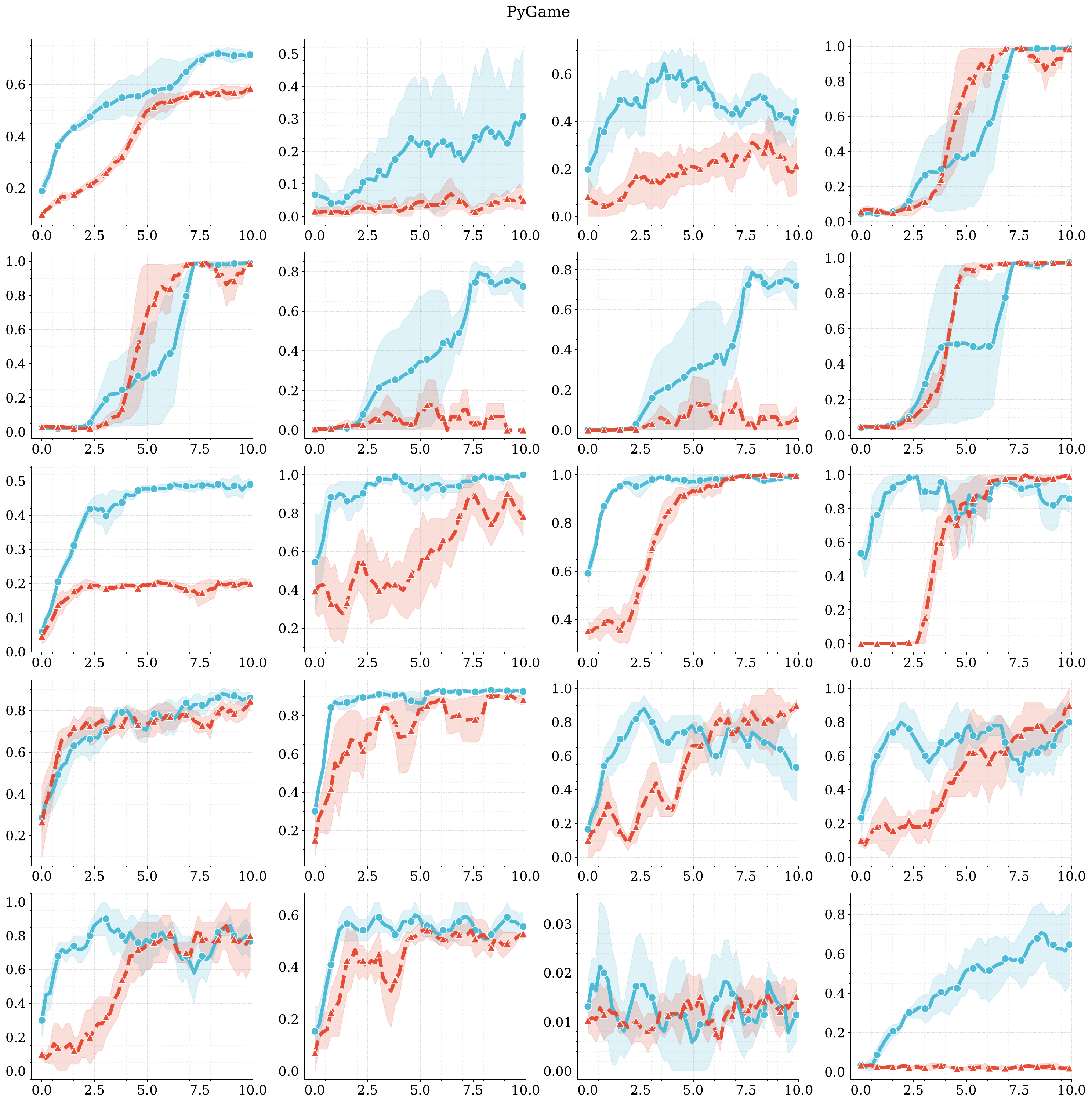}%
\vspace{-0.35cm}
\caption{\textbf{PyGame per-game learning performance.} \hlSmall{MR.Q}, a model-free agent augmented with predictive model-based representations, consistently matches or surpasses the world-model-based approach \hlMedium{\textit{Newt}} across PyGame tasks. Shaded regions denote 95\% confidence intervals (CIs).}
\vspace{-0.3cm}
\label{fig:pygame_per_games}
\end{figure*}

\begin{figure*}[!h]
\centering
\includegraphics[width=\textwidth]{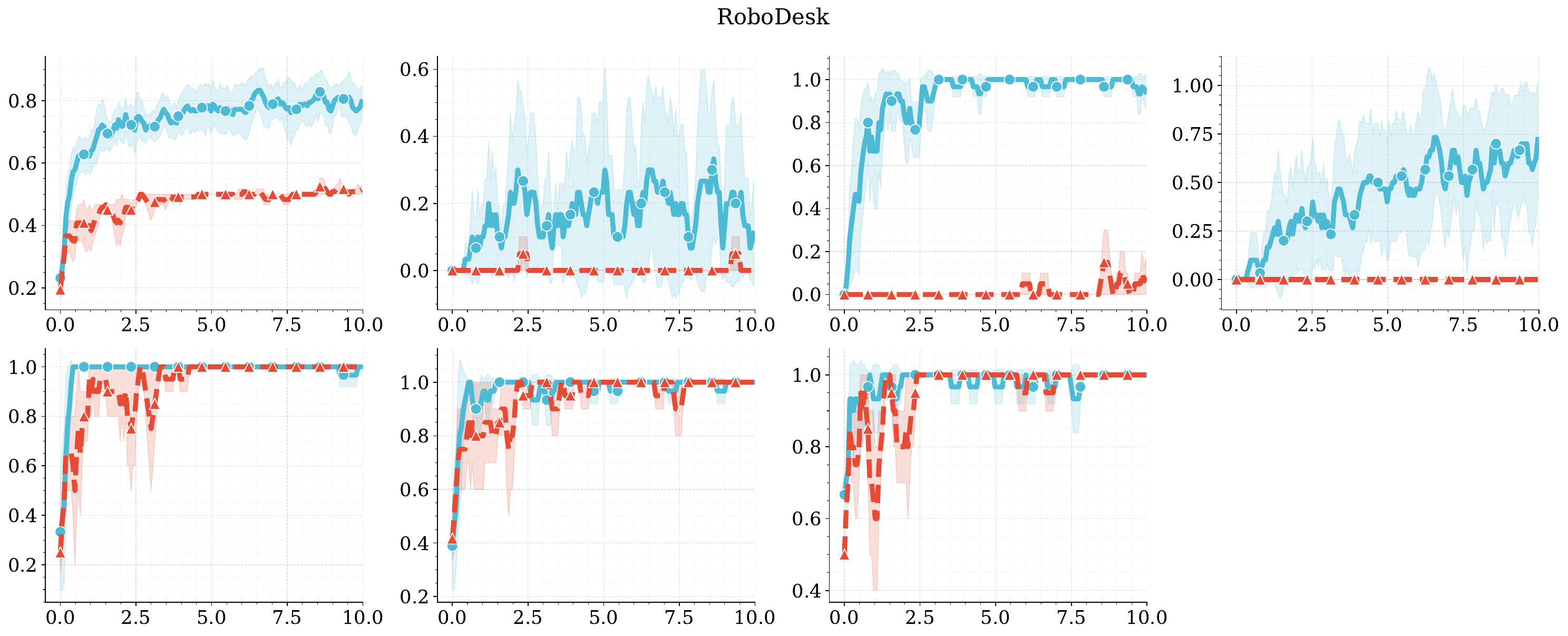}%
\vspace{-0.35cm}
\caption{\textbf{RoboDesk per-game learning performance.} \hlSmall{MR.Q}, a model-free agent augmented with predictive model-based representations, consistently matches or surpasses the world-model-based approach \hlMedium{\textit{Newt}} across RoboDesk tasks. Shaded regions denote 95\% confidence intervals (CIs).}
\vspace{-0.3cm}
\label{fig:robodesk_per_games}
\end{figure*}

\newpage
\clearpage

\section{Finetuning: Per-tasks learning curves}
To evaluate transfer to unseen tasks, we finetune pretrained multitask checkpoints on held-out environments using online RL. All experiments are initialized from the same multitask checkpoint and finetuned under identical interaction budgets. These experiments evaluate whether the representations learned during multitask pretraining transfer effectively to novel tasks and support rapid adaptation under limited additional experience. See \autoref{sec:eval_sacale} for more details.

\begin{figure*}[!h]
    \centering
    \includegraphics[width=0.79\textwidth]{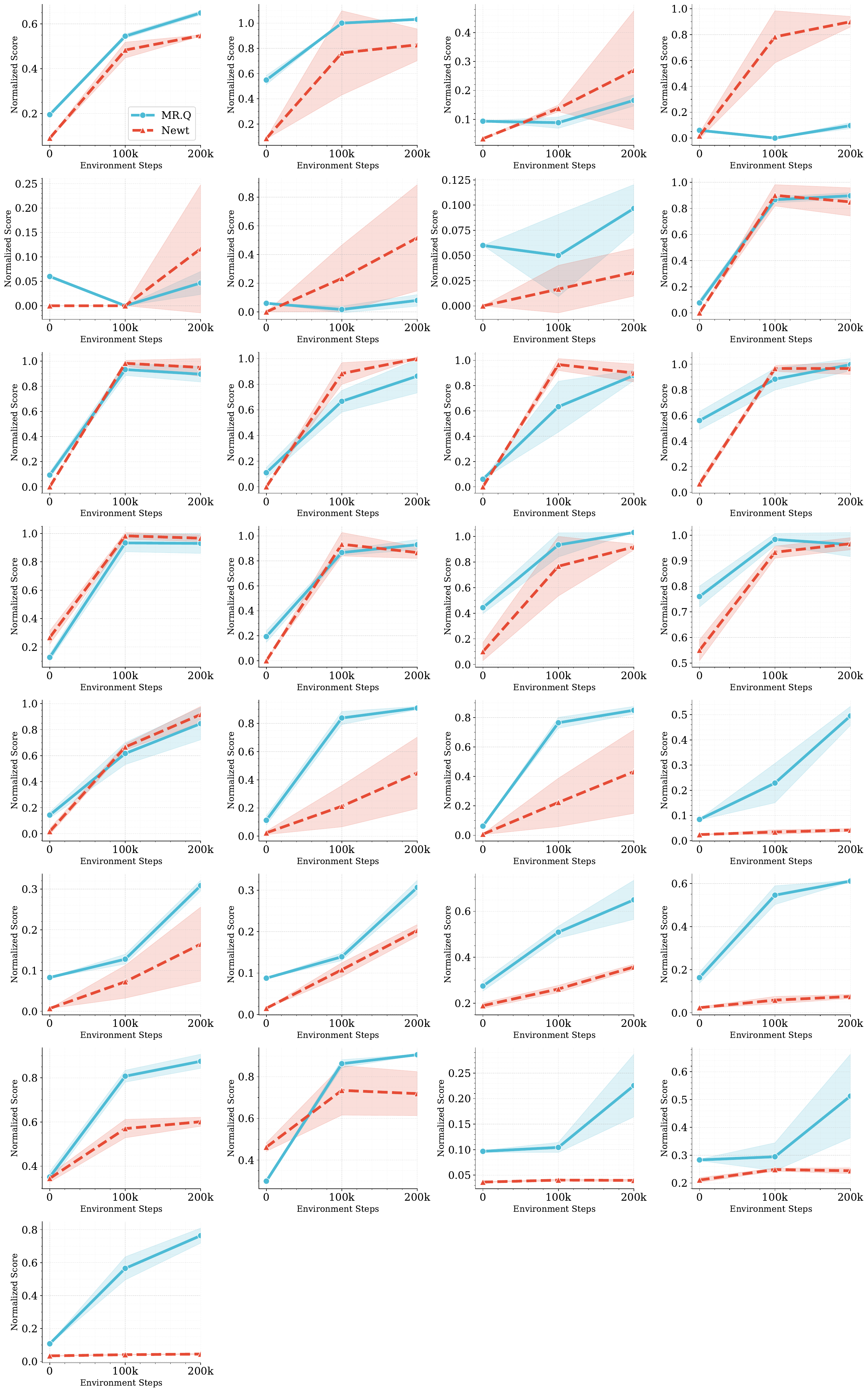}%
    \vspace{-0.35cm}
    \caption{\textbf{Per-task finetuning performance on held-out environments.} Learning curves during online finetuning from pretrained multitask checkpoints. \hlSmall{MR.Q} consistently achieves stronger zero-shot initialization and faster adaptation across the majority of held-out tasks, indicating improved transfer and representation reuse. Shaded regions denote 95\% confidence intervals (CIs).}
    \vspace{-0.3cm}
    \label{fig:atari_per_games}
\end{figure*}

\end{document}